\documentclass[conference]{IEEEtran}
\IEEEoverridecommandlockouts
\usepackage{cite}
\usepackage{amsmath,amssymb,amsfonts}
\usepackage{algorithmic}
\usepackage{graphicx}
\usepackage{textcomp}
\usepackage{makecell}
\usepackage{multirow,bigdelim,dcolumn,booktabs}
\usepackage{array}
\usepackage{float}
\usepackage{enumitem} 
\usepackage[caption=false]{subfig}
\usepackage{url}

\usepackage{etoolbox}
\makeatletter
\patchcmd{\@makecaption}
  {\scshape}
  {}
  {}
  {}
\makeatother

\usepackage{xcolor}
\def\BibTeX{{\rm B\kern-.05em{\sc i\kern-.025em b}\kern-.08em
    T\kern-.1667em\lower.7ex\hbox{E}\kern-.125emX}}
\begin{document}

\title{Empirical Quantitative Analysis of COVID-19 Forecasting Models\\
}

\author{Anonymous}

\author{\IEEEauthorblockN{Yun Zhao\textsuperscript{*}\IEEEauthorrefmark{2}, Yuqing Wang\textsuperscript{*}\IEEEauthorrefmark{2},
Junfeng Liu\IEEEauthorrefmark{2}, Haotian Xia\IEEEauthorrefmark{2}, Zhenni Xu\IEEEauthorrefmark{2}, Qinghang Hong\IEEEauthorrefmark{2}, Zhiyang Zhou\IEEEauthorrefmark{3}, Linda Petzold\IEEEauthorrefmark{2}}
\IEEEauthorblockA{\IEEEauthorrefmark{2}Department of Computer Science, University of California, Santa Barbara, CA, USA\\
\IEEEauthorrefmark{3}Department of Preventive Medicine, Northwestern University Feinberg School of Medicine, Chicago, IL, USA\\
yunzhao, wang603@ucsb.edu}
}

\maketitle
\begingroup\renewcommand\thefootnote{*}
\footnotetext{*Equal contribution}
\endgroup

\maketitle

\begin{abstract}
COVID-19 has been a public health emergency of international concern since early 2020. Reliable forecasting is critical to diminish the impact of this disease. To date, a large number of different forecasting models have been proposed, mainly including statistical models, compartmental models, and deep learning models. However, due to various uncertain factors across different regions such as economics and government policy, no forecasting model appears to be the best for all scenarios. In this paper, we perform quantitative analysis of COVID-19 forecasting of confirmed cases and deaths across different regions in the United States with different forecasting horizons, and evaluate the relative impacts of the following three dimensions on the predictive performance (improvement and variation) through different evaluation metrics: model selection, hyperparameter tuning, and the length of time series required for training. We find that if a dimension brings about higher performance gains, if not well-tuned, it may also lead to harsher performance penalties. Furthermore, model selection is the dominant factor in determining the predictive performance. It is responsible for both the largest improvement and the largest variation in performance in all prediction tasks across different regions. While practitioners may perform more complicated time series analysis in practice, they should be able to achieve reasonable results if they have adequate insight into key decisions like model selection. 

\end{abstract}

\begin{IEEEkeywords}
 COVID-19 Pandemic; Time Series Forecasting; SARIMA Model; SEIR-HCD Model; Deep Learning
\end{IEEEkeywords}

\section{Introduction}
The COVID-19 pandemic has turned the world upside down. It has affected every aspect of people's life, posed numerous threats to global health, and overwhelmed the health care systems in a majority of countries around the world. On March 2021, COVID-19 was flagged as a global pandemic by the World Health Organization. As of 15 May 2021, COVID-19 had resulted in more than 32 million confirmed cases in the United States, and 160 million total reported cases worldwide~\cite{dong2020interactive}. Simultaneously, the ongoing pandemic has caused over 585,000 and 3,369,000 deaths in the United States and worldwide respectively~\cite{dong2020interactive}. The pandemic has triggered devastating social and economic impacts all over the world. Nearly half of the world’s 3.3 billion global workforce are at risk of losing their livelihoods.

In fact, the increasing demand for health care has produced large flows of patients, leading to hospital bed shortages and strain situations in hospitals~\cite{singer1983rationing}. Thereby, for COVID-19 and future pandemics, it is crucial to construct methods to forecast the spread of confirmed and death COVID-19 cases accurately, as they can provide guidance for medical institutions to allocate their resources effectively. Policymakers can also benefit from reliable forecasts to carry out appropriate social intervention strategies to slow down its spreading~\cite{ribeiro2020short,anastassopoulou2020data}. Epidemic forecasting has been considered as a challenging task for a long time. The forecasting of COVID-19 is even harder as various constantly changing factors, such as social and cultural differences, intervention policies, healthcare facilities, influence the transmission rate and mortality rate to a large extent.  For COVID-19 forecasting, there are a large number of research works utilizing different kinds of epidemic models, which can be broadly categorized into three groups: traditional statistical analysis models (e.g., Auto Regressive Integrated Moving Average (ARIMA, \cite{kumar2020forecasting,roy2021spatial}) and Seasonal ARIMA (SARIMA, \cite{satpathy2021predicting,koyuncu2021forecasting}), deep learning based models (e.g., Long Short-Term Memory (LSTM, \cite{chimmula2020time}), Transformer~\cite{jin2020inter} and convolutional neural networks~\cite{minaee2020deep}), and compartmental models such as SIR (Suspected-Infected-Recovered, \cite{ellison2020implications}), SEIR (Suspected-Exposed-Infected-Recovered, \cite{he2020seir}) and SEIRD (Suspected-Exposed-Infected-Recovered-Deceased, \cite{jiang2021epidemiological}). Existing COVID-19 forecasting approaches differ substantially in methods, assumptions, forecast horizons and estimated quantities. Furthermore, these forecasting models confront great challenges in predicting varying situations and tasks accurately, since the circumstances in different regions, including economy, government policy and vaccine coverage, differ tremendously from each other. Different models can make very different projections of COVID-19 cases.  This can result in a large amount of criticism, and leave governments and healthcare officials with some very difficult choices for how to carry out appropriate policies~\cite{kreps2020model,eker2020validity}.  

To this end, we experiment with three models from different categories: SARIMA, SEIR-HCD~\cite{unlu2020epidemic} and Transformer-based Attention Crossing Time Series (ACTS, \cite{jin2020inter}), for COVID-19 daily newly confirmed and mortality case prediction across different regions in the United States with forecast horizons of 7-day or 28-day. We perform hyperparameter tuning for each model and use different lengths of historical data for training. We evaluate the predictive performance through commonly used evaluation metrics in time series forecasting, including the Accuracy, Mean Absolute Percentage Error (MAPE), Weighted Absolute Percentage Error (WAPE), Mean Absolute Error (MAE), Mean Squared Error (MSE), Root Mean Squared Error (RMSE), and Root Mean Squared Logarithmic Error (RMSLE),
all specified in Section~\ref{evaluation}.
Our goals are three-fold. First, we wish to quantify the relative impacts of three dimensions, model selection (i.e. SARIMA, SEIR-HCD, and ACTS), hyperparameter tuning, and training time series length (specified in Table~\ref{date_range}), on the performance improvement and variation across different regions. Second, we seek to understand the relationship between the predictive performance and performance variation caused by each dimension. Third, we want to know which dimensions have larger influence on the performance, such that practitioners are able to pay more attention to those key factors when performing time series analysis in practice. Our experimental results suggest that model selection is the dominant factor that contributes most to both the performance gains and penalties. Furthermore, there is a positive correlation between predictive performance and performance variation. That is, dimensions that bring about higher performance also run the risk of greater performance loss. 

The main contributions of this paper are highlighted as follows:

\begin{enumerate}[label=(\arabic*)]
    \item To the best of our knowledge, this paper is the first effort to conduct a thorough empirical analysis quantifying the predictive performance of time series forecasting of COVID-19.
    \item Our experimental results indicate that model selection brings about the most performance improvement and variation in all forecasting tasks throughout different regions. Furthermore, performance improvement and performance variation of each dimension are clearly positively correlated. In other words, a dimension that brings a larger performance improvement also results in a larger performance variation.
    \item We provide general guidance for practitioners in time series forecasting through quantitative analysis of different dimensions. In particular, our results can guide practitioners regarding which dimension should be prioritized and to be cautious about the risk-benefit trade-off between performance variation and performance improvement from each dimension.
\end{enumerate}

The remainder of this paper is organized as follows.
Section~\ref{related_work} describes related work. The three models and evaluation metrics we use are described in Section~\ref{methods}. Section~\ref{data_exp} presents the datasets and experimental settings. Empirical results are discussed in Section~\ref{results}. Finally, our conclusions are presented in Section~\ref{discussion}. 


\section{Related Work}\label{related_work}
\subsection{Models}

With the emergence and spread of COVID-19, several scientific domains around the world are facing huge research challenges to slow down or arrest the increasing trends of the spread of this disease. Hence, in order to better understand and manage this epidemic, various modeling, estimation, and forecasting methods have been proposed. 

There are a large number of research works utilizing statistical methods to forecast COVID-19 cases (confirmed, recovered and deaths). The ARIMA model is one of the most popular statistical models for times series forecasting, aiming to describe the autocorrelation among time series data. \cite{kumar2020forecasting} employed this model to conduct short-term forecasting for cumulative COVID-19 confirmed, death, and recovered cases on top 15 countries in the world. In~\cite{gupta2020trend}, the ARIMA model and exponential smoothing methods were joint applied in analyzing the trends of the COVID-19 outbreak in India. 
As an extension of ARIMA, SARIMA is capable of modeling a wide range of seasonal data.
It was used to forecast the cumulative COVID-19 cases in top 16 countries~\cite{arunkumar2021forecasting}. Also, it helped \cite{satpathy2021predicting} predict mortality rates of COVID-19 patients.

Other studies apply mathematical models to simulate the epidemics. Epidemiological models that divide the entire population into  different compartments are called compartmental models, which utilize differential
equations to simulate the disease transmission process. The most commonly used ones among compartmental models are the SIR and SEIR models, which are used to analyze the spread of COVID-19. In~\cite{he2020seir}, the SEIR model was introduced to simulate the dynamics of COVID-19. In~\cite{boudrioua2020predicting}, the SIR model was applied to predict the daily infected cases in Algeria. Due to special features of COVID-19 such as its relatively long incubation period, and the high dependency of epidemic trends on artificial factors (e.g., medical resources and quarantine measures), many researchers have proposed the extension of the above two models to better adapt to the characteristics of COVID-19. StochSS Live!~\cite{jiang2021epidemiological} performed inference with Approximate Bayesian Computation algorithms and simulated the COVID-19 cases in two U.S. counties based on the SEIRD model. In~\cite{zou2020epidemic}, the authors proposed a variant of the SEIR model by taking into account the untested/unreported cases of COVID-19. The SEIR-HCD model has been proposed to extend the SEIR model according to characteristics of COVID-19 by adding three additional compartments: H (Hospitalized), C (Critical) and D (Dead)~\cite{unlu2020epidemic}. It was employed to analyze the spread of COVID-19 in France.

For deep learning based approaches, an LSTM-based model was used to forecast the COVID-19 transmission in Canada, Italy, and the United States~\cite{chimmula2020time}. DeepCovid~\cite{rodriguez2020deepcovid} incorporates deep learning and temporal correlations between consecutive forecasts to perform short-term forecasting. In~\cite{hu2020artificial}, a stacked auto-encoder model is proposed to fit the transmission dynamics of the
epidemics and applied to real-time forecasting on confirmed cases in China. ACTS~\cite{jin2020inter} applies detrending and leverages inter-series attention mechanisms on embeddings of time-series segments to obtain the predictions from different regions in the United States.

The models mentioned above achieve their best performance in different situations. They also have their own limitations such as the assumption of linear pattern of time series data in SARIMA models, fixed transmission rate in compartmental models, and lack of interpretability in deep learning approaches. Due to these restrictions, this paper is aimed at presenting a comprehensive study using three models from each category, and taking other factors into consideration such as available historical time series data for training and hyperparameter optimization. Essentially, three models namely SARIMA, SEIR-HCD, and ACTS are applied to forecast the time series of the number of newly confirmed and death cases in the United States across different regions.

\subsection{Performance Metrics}
Performance metrics are essential for evaluating how well the model predictions fit the data. Choosing an appropriate metric for different tasks is crucial for establishing robust and useful models. Commonly used performance measures for time series forecasting include the Accuracy, MAPE, WAPE, MAE, MSE, RMSE, and RMSLE. Each metric has its own strengths and weaknesses in practice. For instance, the benefit of using RMSLE, MAE, MAPE, and WAPE as statistical indicators is that they are more robust to outliers than other metrics. The MSE and RMSE tend to penalize large prediction errors harshly and they are influenced a great deal by extreme values.
Accuracy can provide people with the most intuitive feeling about how close is the predicted value to the actual value. Lower MSE, RMSE, MAE, WAPE, MAPE or RMSLE values, and Accuracy closer to 1 represent more accurate forecasting performances. Due to different characteristics among these metrics, it is tough to 
use a single metric to determine the quality of the model. Hence, we utilize all of these seven metrics to evaluate the predictive performance of three models mentioned above on COVID-19 forecasting of confirmed and death cases.

\section{Methods} \label{methods}
\subsection{SARIMA Model}
ARIMA is one of the most widely used approaches for time series forecasting. Specifically, it makes the prediction by utilizing the lags of time series and lagged forecast errors. An ARIMA model combines the differencing with an autoregression (AR) and a moving average (MA) model. It is characterized by three parameters: $p$, $d$ and $q$, where $p$ is the order of AR model and represents the number of time lags; $d$ is the number of nonseasonal differences required to make the data stationary; $q$ is the order of MA model and represents the number of lagged forecast errors. Altogether, using the backward shift operator $B$, the ARIMA model can be written as:
\begin{align*}
    \phi_p(B) (1-B)^d y_t = \theta_q(B) \varepsilon_t,
\end{align*}
where
\begin{align*}
    \phi_p(B) &= 1 - \phi_1(B) - \phi_2(B^2) - \cdots - \phi_p(B^p), \\
    \theta_q(B) &= 1 + \theta_1(B) + \theta_2 (B^2) + \cdots + \theta_q (B^q).
\end{align*}
Here, $y_t$ is the trajectory value at time $t$, 
$\varepsilon_t$ is normally distributed with zero mean,
and $\phi_i$'s ($i=1,2,\cdots,p$) and $\theta_j$'s ($j=1,2,\cdots,q$) are all unknown scalars.

As an extension of ARIMA, SARIMA admits seasonal components.
Taking seasonality into account,
SARIMA contains non-seasonal ARIMA parameters $p$, $d$, and $q$ and seasonal ones $s$, $P$, $D$, and $Q$. 
Specifically, the SARIMA model is defined as:
\begin{align*}
   \phi_p(B)\Phi_P(B^s)(1-B)^d(1-B^s)^Dy_t = \theta_q(B)\Theta_Q(B^s) \varepsilon_t,
\end{align*}
where
\begin{align*}
    \phi_p(B) &= 1 - \phi_1(B) - \phi_2 (B^2) - \cdots - \phi_p (B^p), \\
    \Phi_P(B^s) &= 1 - \Phi_1(B^s) - \Phi_2(B^{2s}) - \cdots - \Phi_P(B^{Ps}), \\
    \theta_q(B) &= 1 + \theta_1(B) + \theta_2(B^2) + \cdots + \theta_q(B^q), \\
    \Theta_Q(B^s) &= 1 + \Theta_1(B^s) + \Theta_2(B^{2s}) + \cdots + \Theta_Q(B^{Qs}).
\end{align*}
Here, $\Phi_i$'s ($i=1,2,\cdots,P$) and $\Theta_j$'s ($j=1,2,\cdots,Q$) are parameters to be estimated;
$P$, $D$, and $Q$ are seasonal counterparts of $p$, $d$, and $q$, respectively;
$s$ is the time-length of a single seasonal period. In our context, $s=7$ (days).

The statistical measures, including Accuracy, MAPE, WAPE, MAE, MSE, RMSE, and RMSLE, are used to evaluate the predictive performance of the SARIMA model.

\subsection{SEIR-HCD Model}
\begin{figure}[ht]
\centering
  \includegraphics[width=\linewidth]{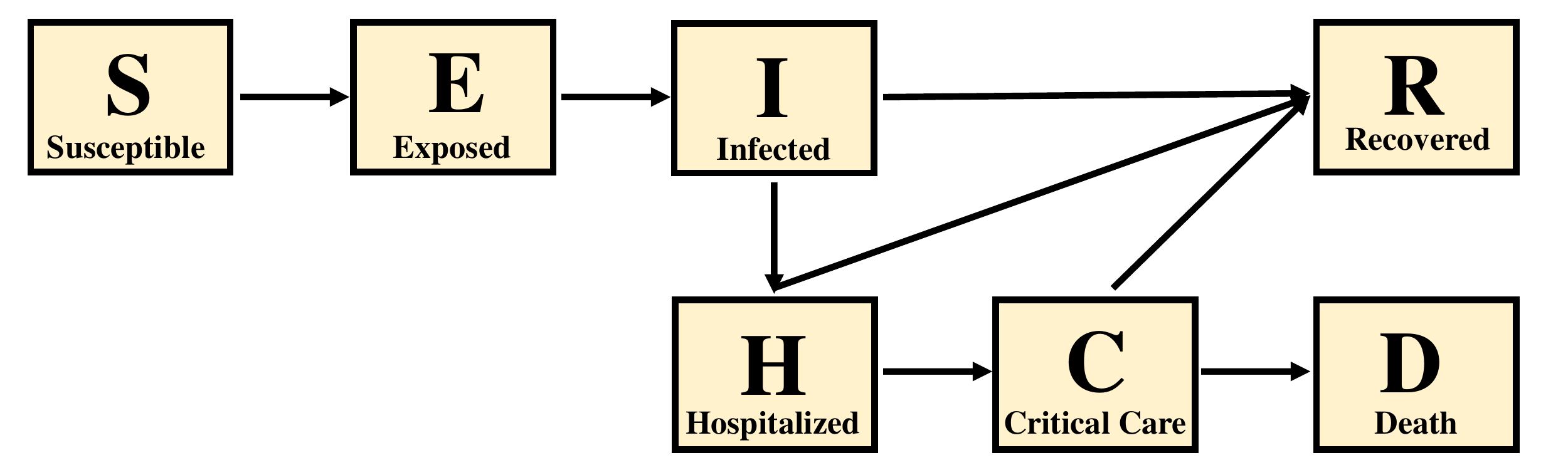}
  \caption{Disease transmission flow of the SEIR-HCD model. Infected individuals may be hospitalized after a period of time. A proportion of infected agents switches to critical state, which requires intensive care; while the rest recovers. In critical cases, a certain proportion of people eventually dies.}
  \label{SEIRHCD_flowchart}
\end{figure}

Classic epidemiological compartmental models such as SIR and SEIR have been widely applied to simulate the spread of diseases in a population. Since COVID-19 has a relatively long incubation period (5--14 days), during which there may be carriers who do not show any symptoms of the disease, we use an extended SEIR model, namely SEIR-HCD, with three additional compartments, which considers seven population compartments: susceptible (S), exposed (E), infected (I), recovered (R), hospitalized (H), critical care (C), and death (D). The disease transmission flow of the model is sketched in Fig.~\ref{SEIRHCD_flowchart}. 

The model is comprised of a system of ordinary differential equations (ODE):
\begin{align*}
 \frac{dS}{dt} &= - \frac{R_0 IS}{t_{\inf}}, \\
  \frac{dE}{dt} &= \frac{R_0 IS}{t_{\inf}} - \frac{E}{t_{\rm inc}}, \\
  \frac{dI}{dt} &= \frac{E}{t_{\rm inc}} - \frac{I}{t_{\inf}}, \\
  \frac{dR}{dt} &=  \frac{\beta I}{t_{\inf}} + \frac{(1-\gamma)H}{t_{\rm hosp}}, \\
  \frac{dH}{dt} &= \frac{(1-\beta)I}{t_{\inf}} + \frac{(1-\delta)C}{t_{\rm crt}} - \frac{H}{t_{\rm hosp}},\\
  \frac{dC}{dt} &= \frac{\gamma H}{t_{\rm hosp}} - \frac{C}{t_{\rm crt}}, \\
  \frac{dD}{dt} &= \frac{\delta C}{t_{\rm crt}},
\end{align*}
with non-negative initial conditions $S(0) = S_0$, $E(0) = E_0$, $I(0) = I_0$, $R(0) = R_0$, $H(0) = H_0$, $C(0) = C_0$, and $D(0) = D_0$. $R_0$ is the basic reproduction number for the coronavirus (i.e. the number of secondary infections each infected individual produces), $t_{\inf}$ is the average infectious period of COVID-19, $t_{\rm inc}$ is the average incubation period of an infected agent, $t_{\rm hosp}$ is the average hospitalized period (i.e. average length of hospital stay before the patient recovers or becomes critical), and $t_{\rm crt}$ is the average critical period (i.e. average time for a hospitalized patient to enter into a critical state since initial check-in).
$\beta$, $\gamma$, and $\delta$ refer to ratios of asymptomatic infected individuals, hospitalized patients who switched to a critical state, and critical patients that result in fatalities, respectively.

We employ the L-BFGS-B optimization algorithm~\cite{zhu1997algorithm,virtanen2020scipy} and optimize the model by finding the above $8$ parameters: $R_0$, $t_{\inf}$, $t_{\rm inc}$, $t_{\rm hosp}$, $t_{\rm crt}$, $\beta$, $\gamma$, and $\delta$. 

The statistical measures, including Accuracy, MAPE, WAPE, MAE, MSE, RMSE, and RMSLE, are used to evaluate the SEIR-HCD model performance.

\subsection{Transformer-based Model (ACTS)}
The Transformer model has been proven to have great potentials for time series forecasting~\cite{wu2020deep,li2019enhancing}. The ACTS model~\cite{jin2020inter} is a new neural forecasting model based on Transformer that performs forecasts by comparing and utilizing similar patterns across time series detected from different geographic regions. It consists of three major components: detrending, attention module, and joint training.

\paragraph{Detrending} ACTS employs a learnable Holt smoothing model to detrend long-term trends of the raw time series and leave the remaining time series (i.e. the residual). Linear extrapolation is used to generate forecasts based on long-term trends. The residual time series are then fed to the following attention module.

\paragraph{Attention module} The attention module is composed of two components: embeddings and inter-series attention. The attention mechanism investigates the relationship among different regions that have been undergoing the pandemic.

\textbf{Embeddings.} The residual time series after detrending is normalized by min-max normalization, which can be considered as a way of smoothing. Consequently, the first and last values of the normalized time series will always be $0$ and $1$, respectively. Then, a convolution layer is applied to encode the normalized time series into segment features, followed by an average pooling layer (segment embeddings) to model the similarity in different regions at different time periods. Likewise, another convolution-pooling layer is employed to encode the following incidents over $H$ days (i.e. forecasting horizon) after each segment into development embedding. It represents the succeeding development after encoded segments and will serve as references for the target region forecasting.  

\textbf{Inter-series attention.}
Following the embeddings, the dot-product attention is used to compute and combine the values of segments. Specifically, segment embeddings are linearly mapped to query vectors $q_{t}^{i}$ and key vectors $k_{i}^{t}$ and development embeddings are projected to value vectors $v_{t}^{i}$. The equations are given by:
\begin{align*}
&q_{t}^{i} = W_{Q}p_{t}^{i},\\
&k_{t}^{i} = W_{K}p_{t}^{i},\\
&v_{t}^{i} = W_{V} g_{t}^{i}.
\end{align*}
Then, the model takes $q_{T}^{i_{0}}$, the last segment of target region $i_{0}$, to compute its similarity with keys from all other regions and time segments to obtain a weighted sum of values:
\begin{align*}
    \hat{v}_T^{i_0} = \frac{\sum_{i, t \in \Omega} v_t^i
            \exp( \langle q_{T}^{i_{0}}, k_t^i \rangle)
        }{\sum_{i,t \in \Omega} \exp( \langle q_{T}^{i_{0}}, k_t^i \rangle },
\end{align*}
where $\Omega = [1,N] \times [l, T-H]$, $N$ is the region, $l$ is the length of the time period, and $T$ is the given date. Finally, $\hat{v}_T^{i_0}$ is linearly projected to an estimate and inversely transformed to an estimate $\hat{y}_{T+1:T+H}^{i_{0}}$. To this end, the estimated value is added to the extrapolation from the detrending module to generate the final forecasting result $y_{T+1:T+H}^{i_{0}}$.
        
\paragraph{Joint training} The model is trained by minimizing an aggregation of prediction errors based on historical records in the different regions and time periods.

The statistical measures, including Accuracy, MAPE, WAPE, MAE, MSE, RMSE, and RMSLE, are used to evaluate the performance of the ACTS model.

\subsection{Evaluation Metrics}\label{evaluation}
All of the above forecasting models are evaluated using the following seven performance metrics:
\begin{align*}
    {\rm Accuracy} &= 1 - \sqrt{\frac{1}{n}\sum_{t=1}^n \bigg( \frac{y_t - \hat{y}_t}{y_t + 1}\bigg)^2},\\
    {\rm MAPE} &= \frac{1}{n} \sum_{t=1}^n \bigg| \frac{y_t - \hat{y}_t}{y_t} \bigg| \times 100\%,\\
    {\rm WAPE} &= \frac{\sum_{t=1}^n |y_t - \hat{y}_t| }{\sum_{t=1}^n |y_t|} \times 100\%, \\
    {\rm MAE} &= \frac{1}{n} \sum_{t=1}^n |y_t - \hat{y}_t|, \\
    {\rm MSE} &= \frac{1}{n} \sum_{t=1}^n (y_t - \hat{y}_t)^2, \\
    {\rm RMSE} &= \sqrt{\frac{1}{n} \sum_{t=1}^n (y_t - \hat{y}_t)^2}, \\
    {\rm RMSLE} &= \sqrt{\frac{1}{n}\sum_{t=1}^n\{\ln(\hat{y}_t + 1) - \ln(y_t + 1)\}^2}, 
\end{align*}
where $y_t$ are the true values, $\hat{y}_t$ are the predicted values, $\bar{y}_t = n^{-1}\sum_{t=1}^n y_t$, and $n$ is the testing sample size.

\begin{table*}[htbp]
\centering
\caption{Training date ranges for different forecasting tasks and regions.}
\label{date_range}
\begin{tabular}{cccccc}
\Xhline{1.5pt}
TS Length Description & Applicability &
  \makecell{7-day ahead forecasts \\ on confirmed cases} &
  \makecell{28-day ahead forecasts \\ on confirmed cases} &
  \makecell{7-day ahead forecasts \\ on death cases} &
  \makecell{28-day ahead forecasts \\ on death cases} \\
 \hline
 \makecell{starting date chosen \\ since the appearance \\ of vaccine} & all the states & \makecell{December 15, 2020 - \\May 8, 2021} & \makecell{December 15, 2020 - \\April 17, 2021} &  \makecell{December 15, 2020 - \\May 8, 2021} & \makecell{December 15, 2020 - \\April 17, 2021} \\
 \hline
 \makecell{length of historical \\TS = 200} & all the states & \makecell{October 21, 2020 - \\May 8, 2021} & \makecell{September 30, 2020 - \\April 17, 2021} & \makecell{October 21, 2020 - \\May 8, 2021} & \makecell{September 30, 2020 - \\April 17, 2021} \\
 \hline
\multirow{10}{*}{\makecell{all of the available \\ TS data since the \\ first case appeared \\in each state}} & California & \makecell{January 26, 2020 -\\May 8, 2021} & \makecell{January 26, 2020 -\\April 17, 2021} & \makecell{March 4, 2020 -\\May 8, 2021} & \makecell{March 4, 2020 -\\April 17, 2021} \\ & New York & \makecell{March 3, 2020 -\\May 8, 2021} & \makecell{March 3, 2020 -\\April 17, 2021} & \makecell{March 11, 2020 -\\May 8, 2021} & \makecell{March 11, 2020 -\\April 17, 2021} \\ & Texas & \makecell{March 5, 2020 -\\May 8, 2021} & \makecell{March 5, 2020 -\\April 17, 2021} & \makecell{March 17, 2020 -\\May 8, 2021} & \makecell{March 17, 2020 -\\April 17, 2021} \\ & Minnesota & \makecell{March 6, 2020 -\\May 8, 2021} & \makecell{March 6, 2020 -\\April 17, 2021} & \makecell{March 21, 2020 -\\May 8, 2021} & \makecell{March 21, 2020 -\\April 17, 2021} \\ & Hawaii & \makecell{March 7, 2020 -\\May 8, 2021} & \makecell{March 7, 2020 -\\April 17, 2021} & \makecell{March 24, 2020 -\\May 8, 2021} & \makecell{March 24, 2020 -\\April 17, 2021}\\
\Xhline{1.5pt}
\end{tabular}
\end{table*}
\begin{table*}[ht]
\centering
\caption{Summary statistics for the COVID-19 historical data in CA, NY, TX, MN, and HI. Conf refers to the data on daily confirmed cases.}
\label{data_statistics}
\begin{tabular}{c|ccccccccc}
\Xhline{1.5pt}
        & Min & Max & Mean & Variance & 25th Percentile & Median & 75th Percentile & Skewness & Kurtosis \\
\hline
Conf-CA &  0   &  62168   & 7932.811 & 125727153.293   & 1828.5  & 3529.0 & 8380.0 & 2.311 & 4.907         \\
Death-CA &  0   & 1086    & 143.403  & 25900.787  & 47.0       &  85.0   &  160.0      &   2.146       &   5.004     \\
Conf-NY &  3   & 27644    & 4757.128  & 20224391.370  &  852.25      & 3139.5     &  7767.75      &  1.136        &   1.279      \\
Death-NY &  0   &  1273   & 123.286  & 37885.338  &  16.25     &  59.0     & 141.0       &  3.103        &  10.398     \\
Conf-TX &  1   &  36283   & 6716.482  & 41642526.834 & 1872.25      &  4868.0   & 9047.75       & 1.598         &   2.658     \\
Death-TX &  0   & 700    & 120.241  & 11851.261 &  36.75     & 89.5      &   173.5     & 1.561       &   3.033 \\
Conf-MN &   0  & 9022    & 1364.646   & 2520586.275 &   480.5   &  843.0     &  1565.5      &  2.447       &  6.314   \\
Death-MN &   0  & 140    & 17.729 & 388.714  &  6.0      & 11.0      &   22.0     &   2.342       & 6.558         \\
Conf-HI &   0  & 483    & 81.571 & 6241.035  & 14.25       & 69.0      &   119.075     &  1.386       &  2.534        \\
Death-HI &   0  &  59   & 1.189 &  11.370 & 0.0  &       0.0 & 2.0    &  12.613        &  206.662 \\
\Xhline{1.5pt}
\end{tabular}
\end{table*}

\section{Data and Experiments} \label{data_exp}
The COVID-19 time series data are publicly
available at JHU-CSSE~\cite{dong2020interactive}. We focus on the univariate time series data of daily confirmed and death cases from each of the five states in the United States, including California (CA), New York (NY), Texas (TX), Minnesota (MN), and Hawaii (HI). The dataset we used covers the reports up to May 15, 2021. We performed 7-day ahead and 28-day ahead forecasts on the newly confirmed and death cases of the above five states, given different lengths of historical data for training. The specific historical date ranges for different forecasting tasks in each state are shown in Table~\ref{date_range}. Table~\ref{data_statistics} provides the summary statistics of each univariate time series used in this study (i.e. all the available historical time series data on confirmed cases and deaths for each state). Due to the characteristics of time series data for confirmed and death cases such as high variance and skewed distribution, different data preprocessing pipelines and experimental settings for each model are described below.

\subsection{Experimental Procedures for SARIMA}
The original COVID-19 time series data displays the non-stationary and high-variance behaviors, which can be reflected from the ACF and the PACF plots of daily confirmed and death cases for each state. We apply the Box--Cox transformation to stabilize the variance. Then, we take lag-one difference twice for the confirmed data and lag-one difference once for the death data to remove the trend. The plots of the original training data and the processed data are shown in Table~\ref{org_pro}. Table~\ref{ADF} suggests that all of the processed time series are stationary through the augmented Dickey--Fuller (ADF) test. Next, we apply grid search for hyperparameter tuning of the SARIMA model. The hyperparameter search space is listed in Table~\ref{hyperparameters_SARIMA}. The maximum likelihood estimation (MLE) is employed to fit the model. Diagnostic checks are performed via residual plots and Q-Q plots to 
evaluate the effectiveness of the SARIMA model.  Finally, the inverse Box--Cox transformation is applied to the forecasting results and different metrics are used to compare our results and the reported ground truths.

\begin{table*}[htbp]
\centering
\caption{Original time series data and corresponding processed time series data through Box--Cox transformation and \\ differencing for confirmed and death cases in the state of CA, NY, TX, MN, and HI for SARIMA model.}
\label{org_pro}
\begin{tabular}{ccccc}
\Xhline{1.5pt}

\textbf{State} & \multicolumn{2}{c}{ \bfseries Confirmed Cases} & \multicolumn{2}{c}{\bfseries Death Cases}         \\
      & Original TS     & Processed TS    & Original TS & Processed TS \\
California    &   \begin{minipage}{.15\textwidth}
      \includegraphics[width=1in]{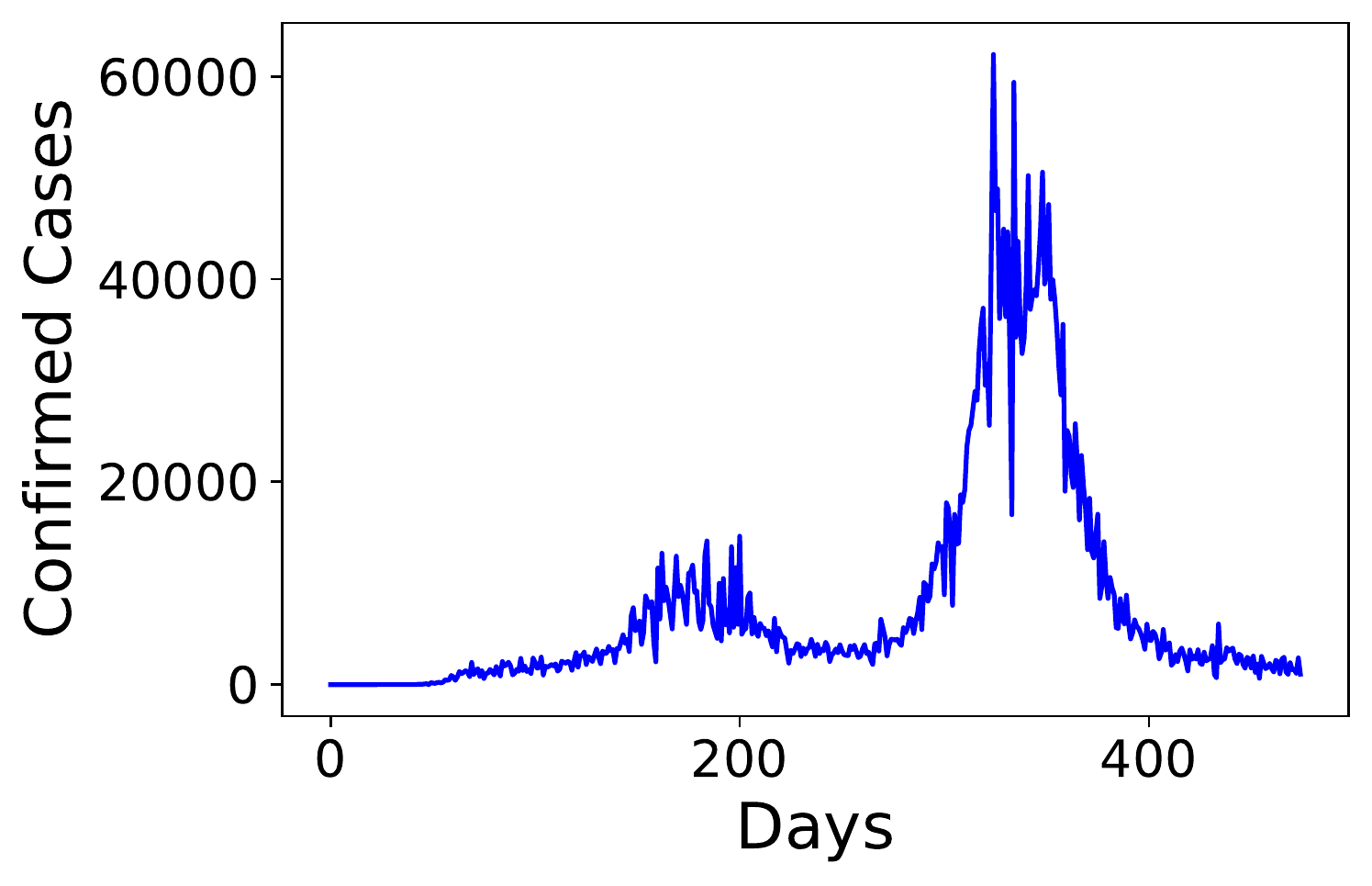}
    \end{minipage} & \begin{minipage}{.15\textwidth}
      \includegraphics[width=1in]{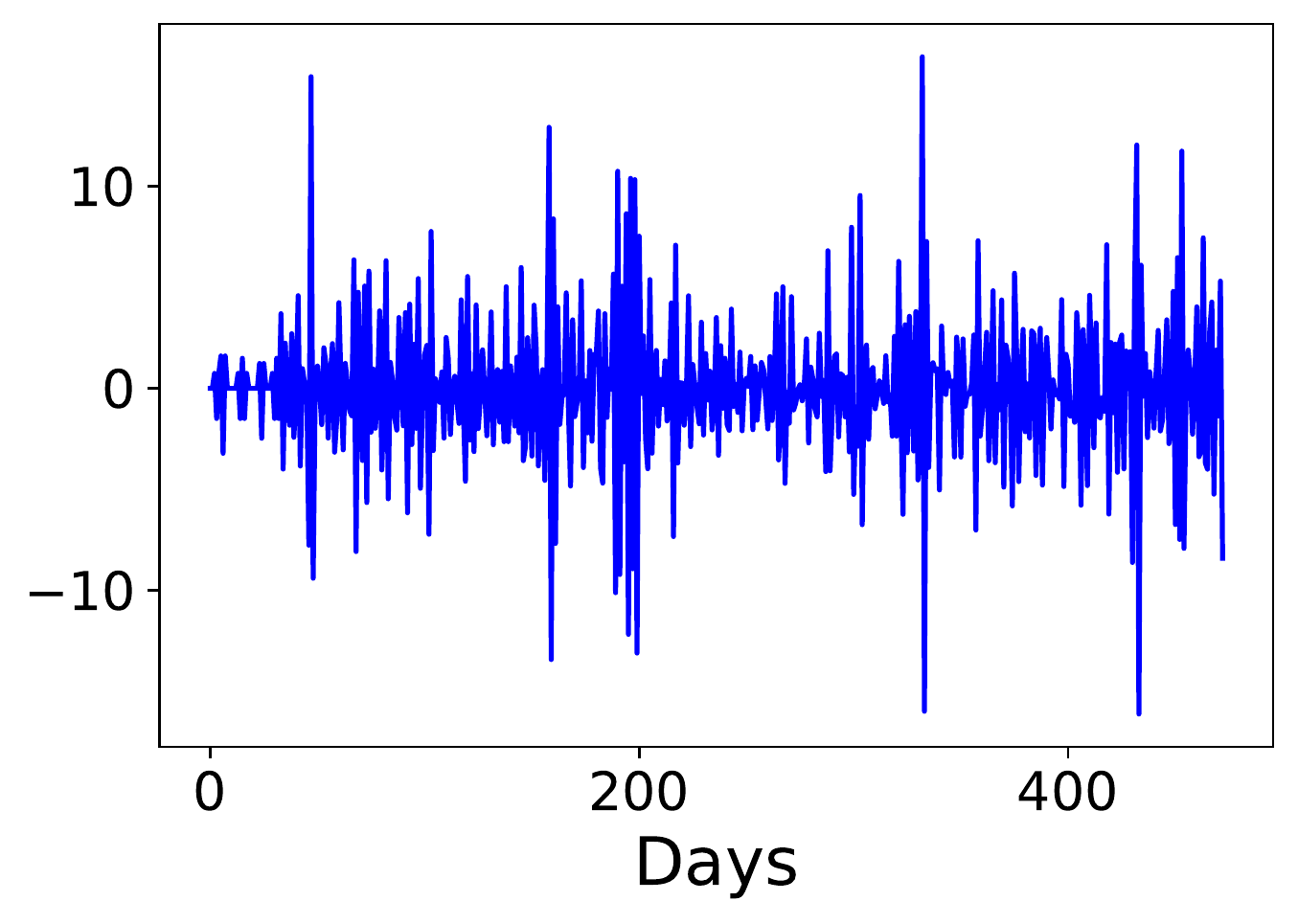}
    \end{minipage}& \begin{minipage}{.15\textwidth}
      \includegraphics[width=1in]{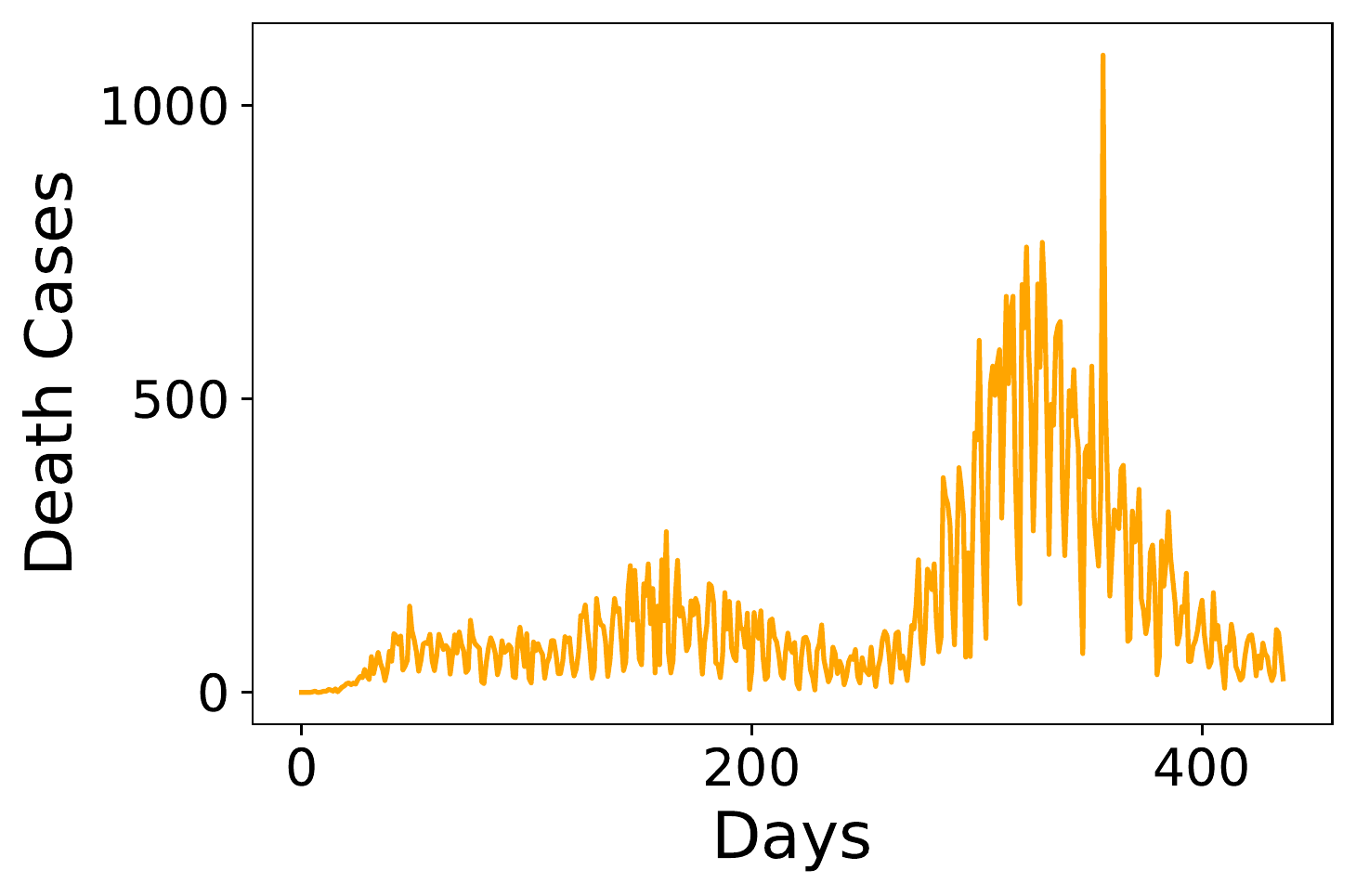}
    \end{minipage}  & \begin{minipage}{.15\textwidth}
      \includegraphics[width=1in]{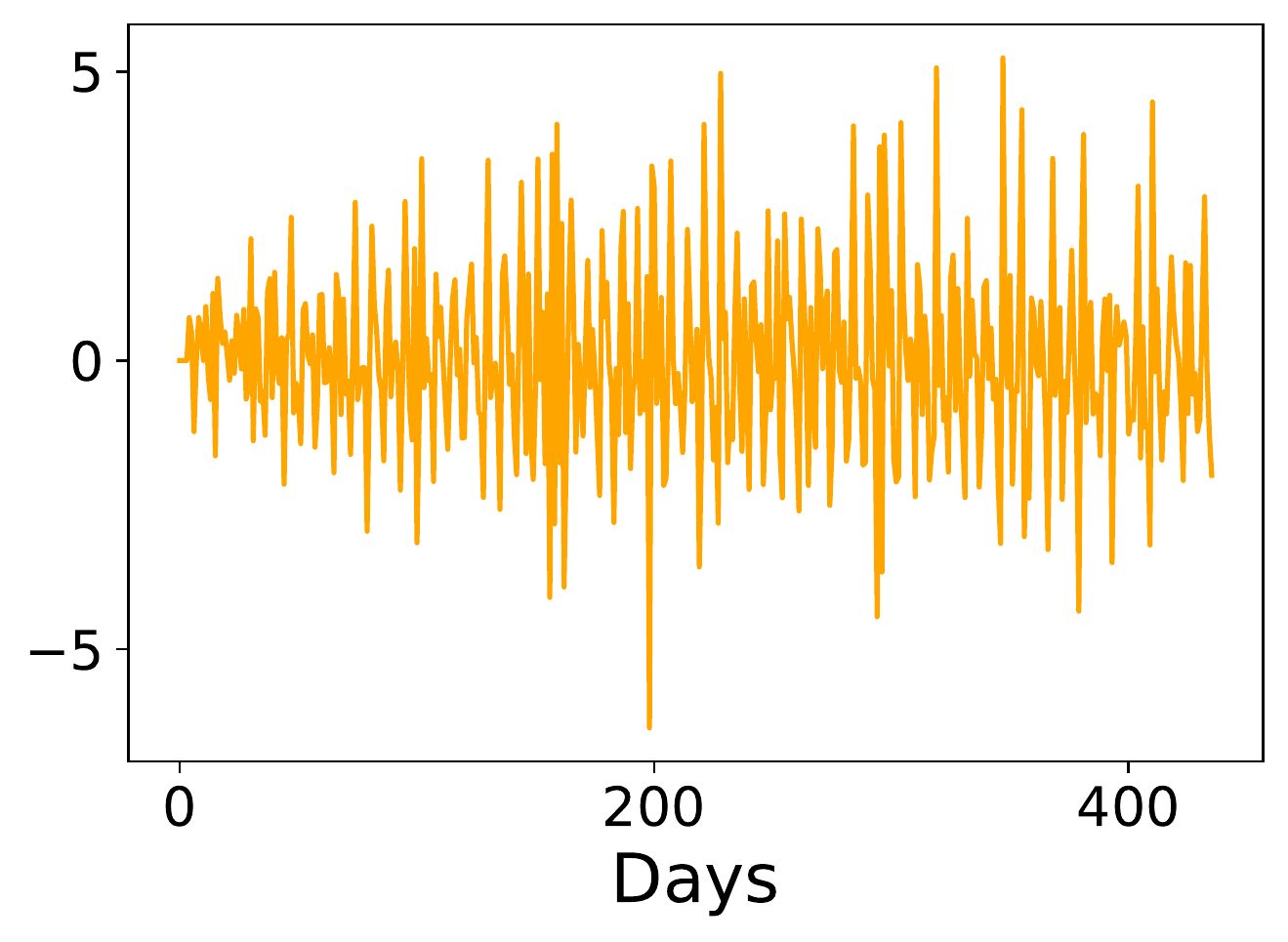}
    \end{minipage} \\
New York    &   \begin{minipage}{.15\textwidth}
      \includegraphics[width=1in]{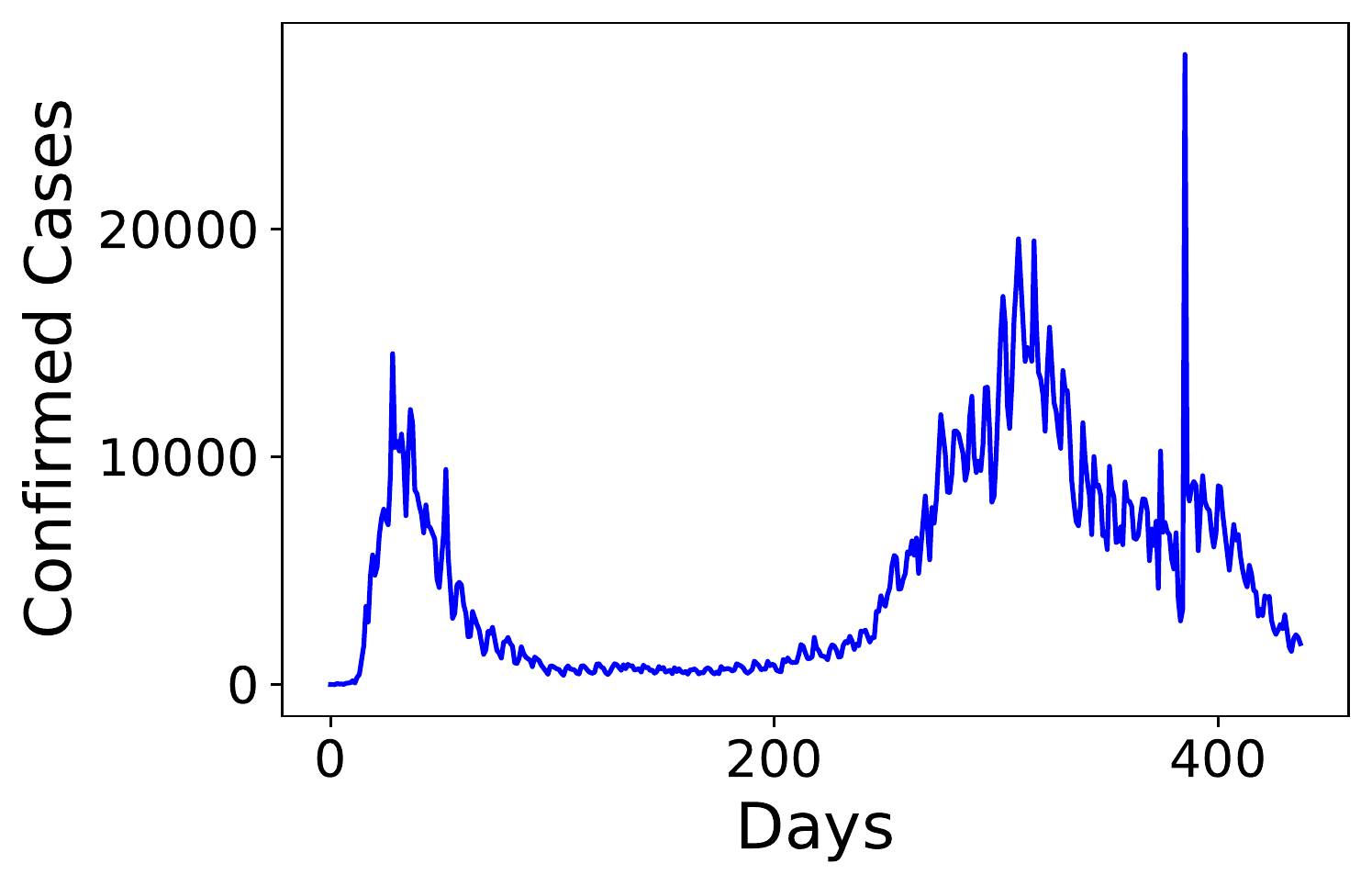}
    \end{minipage} & \begin{minipage}{.15\textwidth}
      \includegraphics[width=1in]{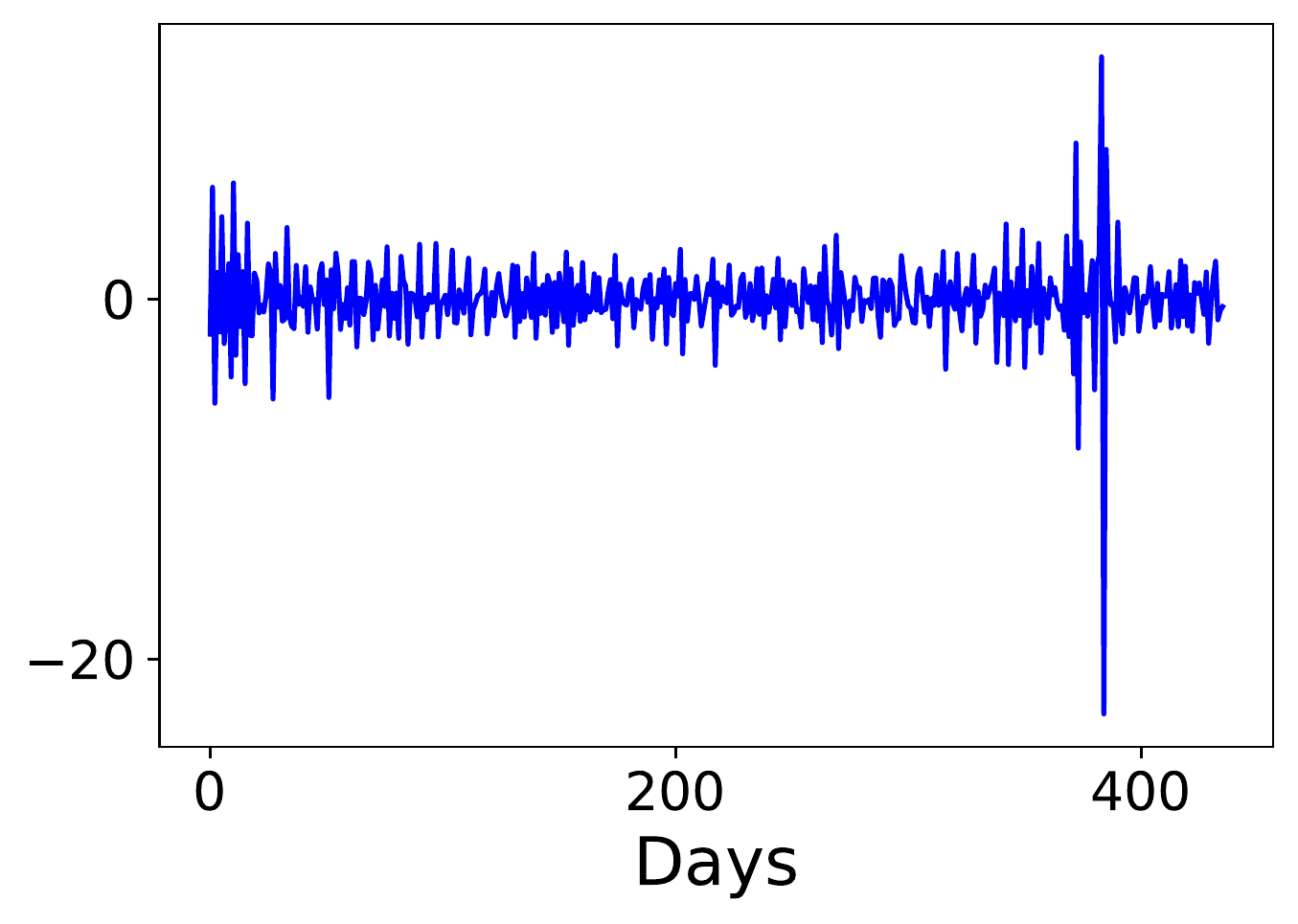}
    \end{minipage}& \begin{minipage}{.15\textwidth}
      \includegraphics[width=1in]{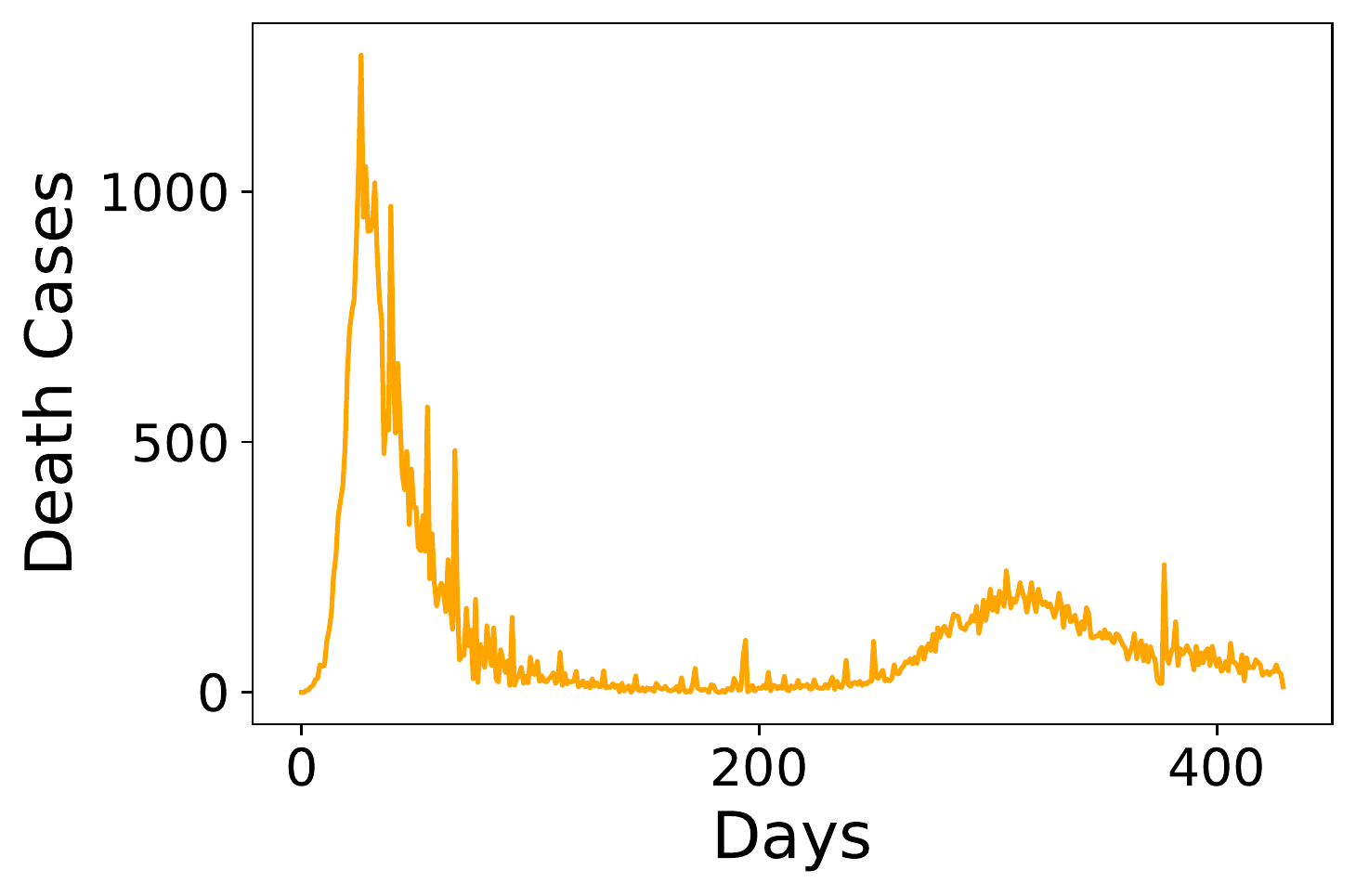}
    \end{minipage}  & \begin{minipage}{.15\textwidth}
      \includegraphics[width=1in]{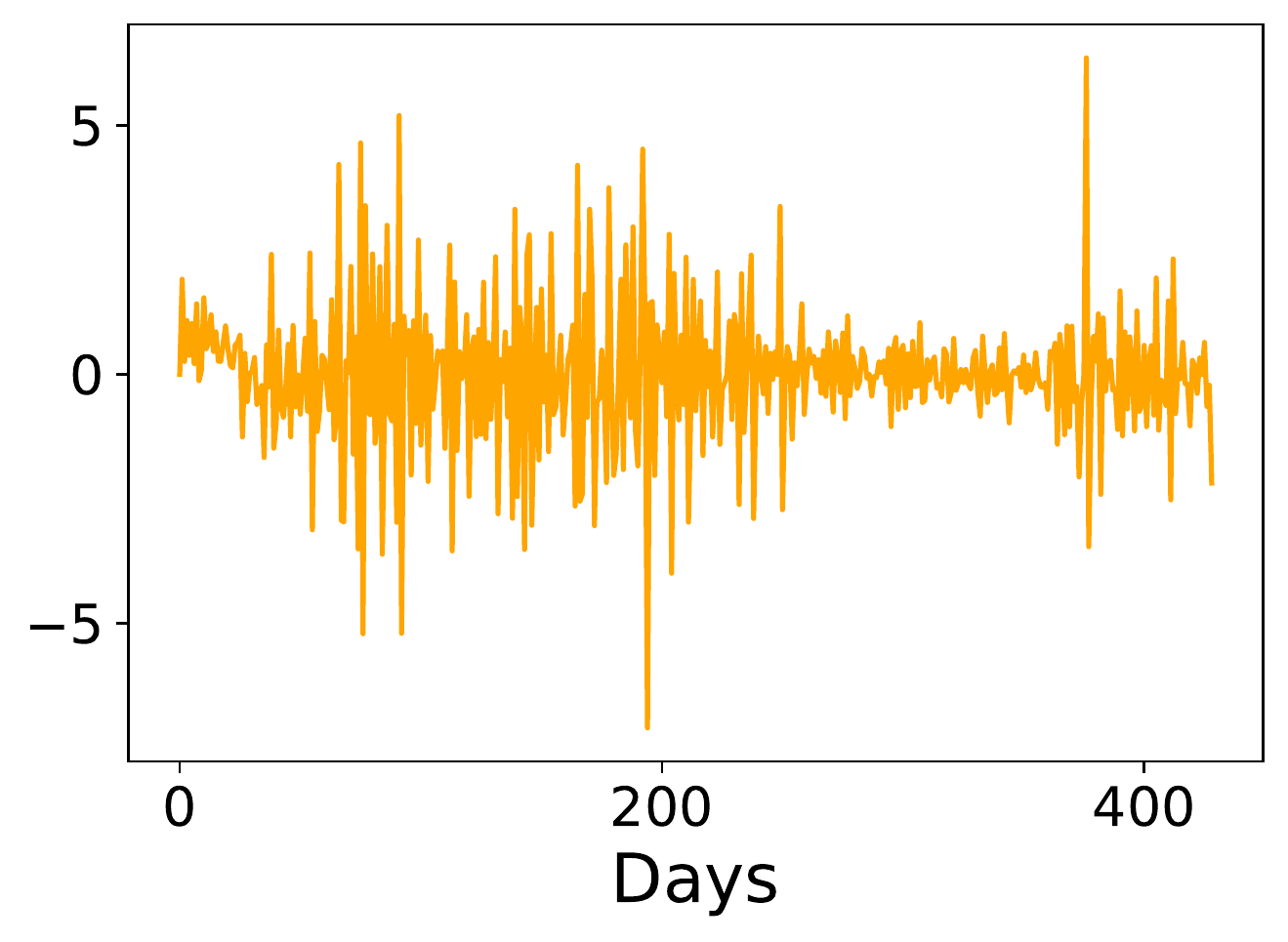}
    \end{minipage}  \\
Texas   &  \begin{minipage}{.15\textwidth}
      \includegraphics[width=1in]{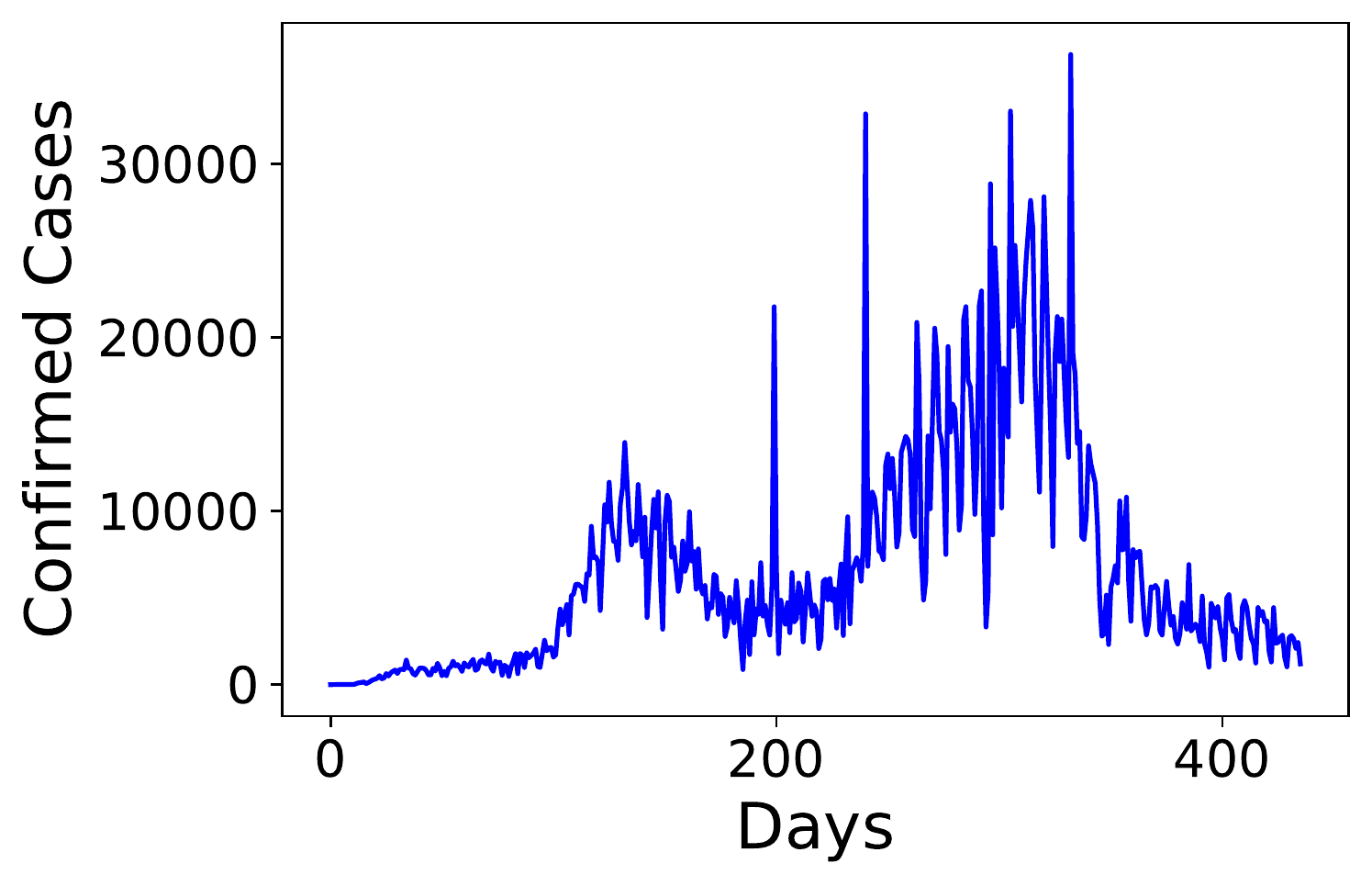}
    \end{minipage} & \begin{minipage}{.15\textwidth}
      \includegraphics[width=1in]{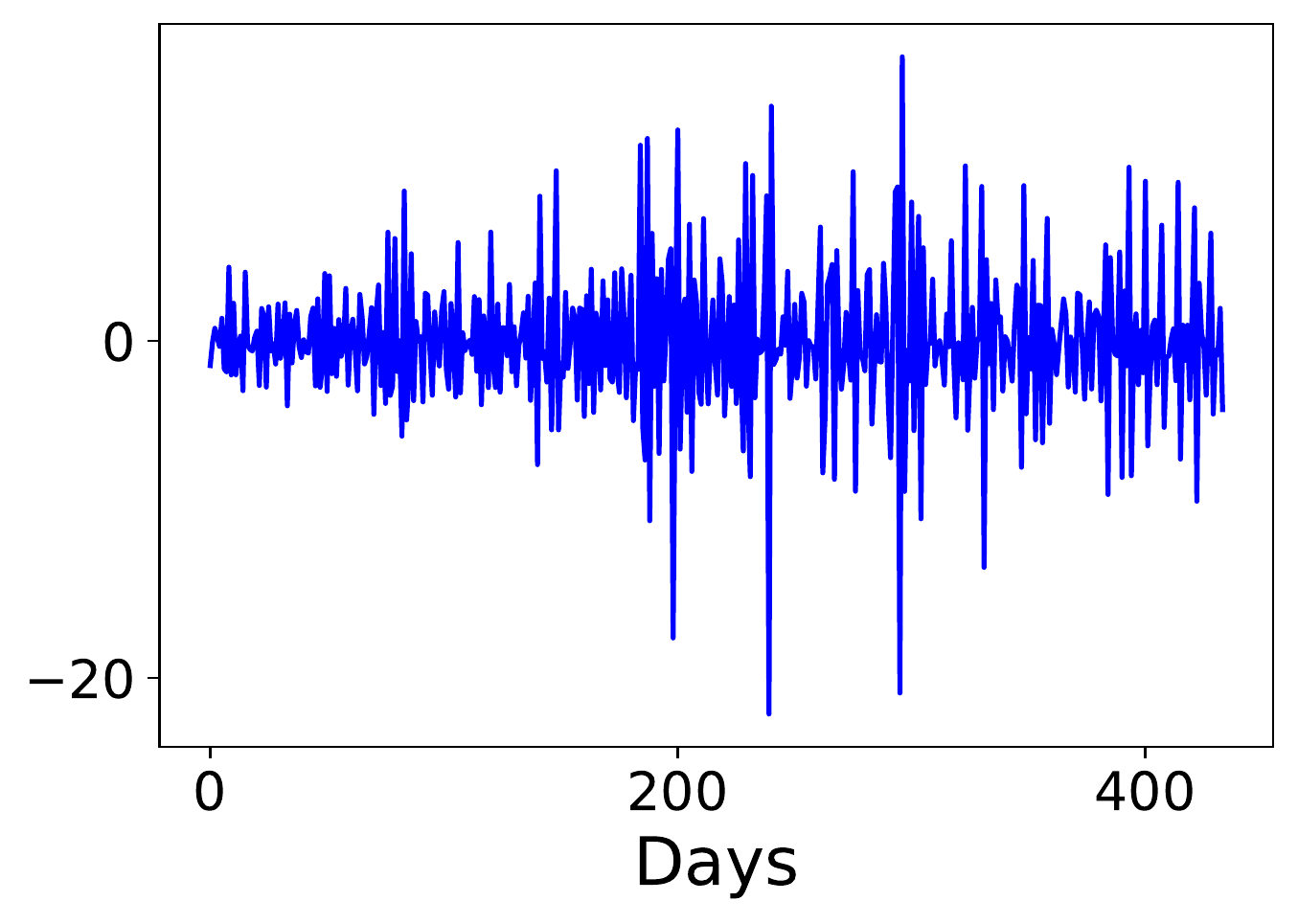}
    \end{minipage}& \begin{minipage}{.15\textwidth}
      \includegraphics[width=1in]{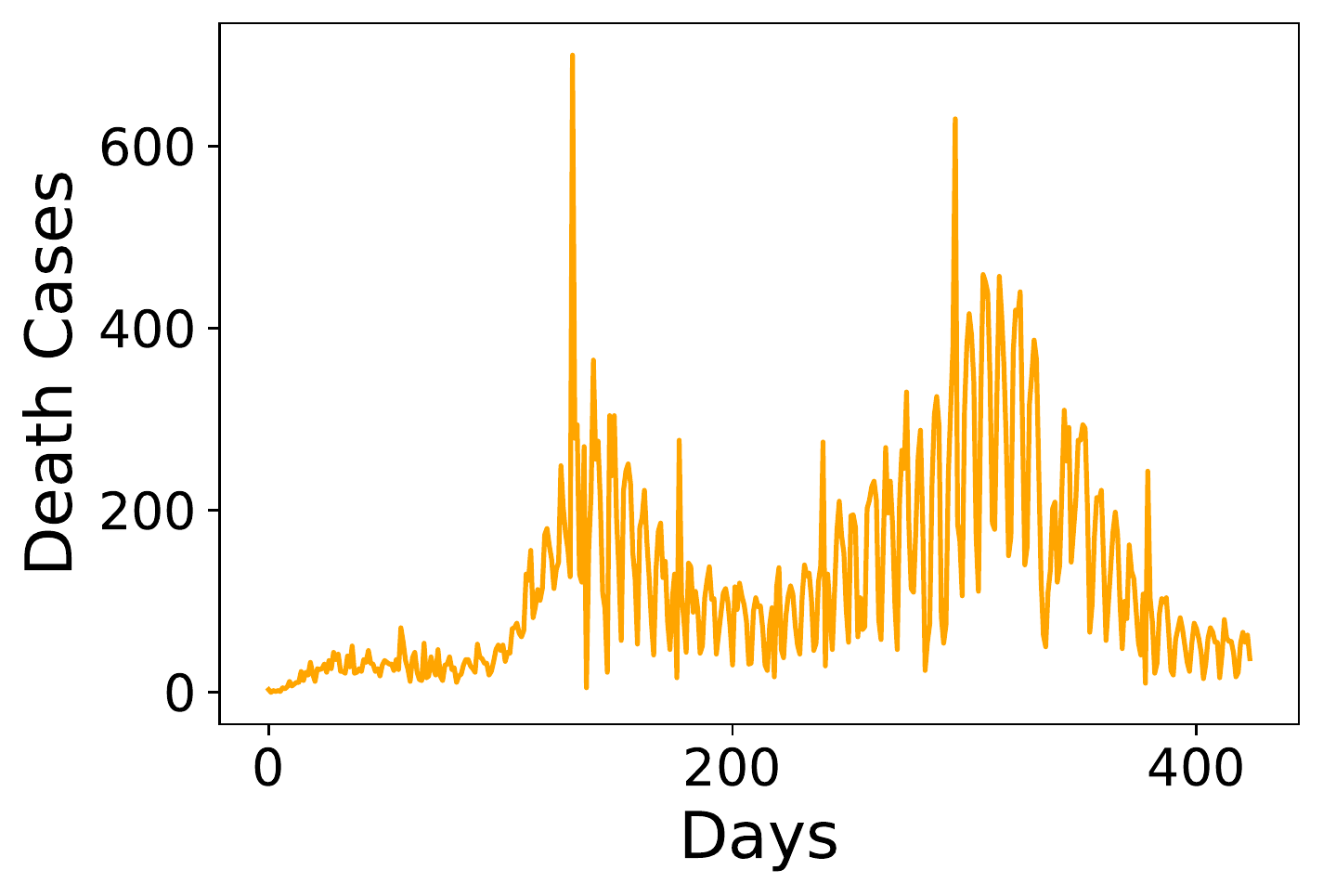}
    \end{minipage}  & \begin{minipage}{.15\textwidth}
      \includegraphics[width=1in]{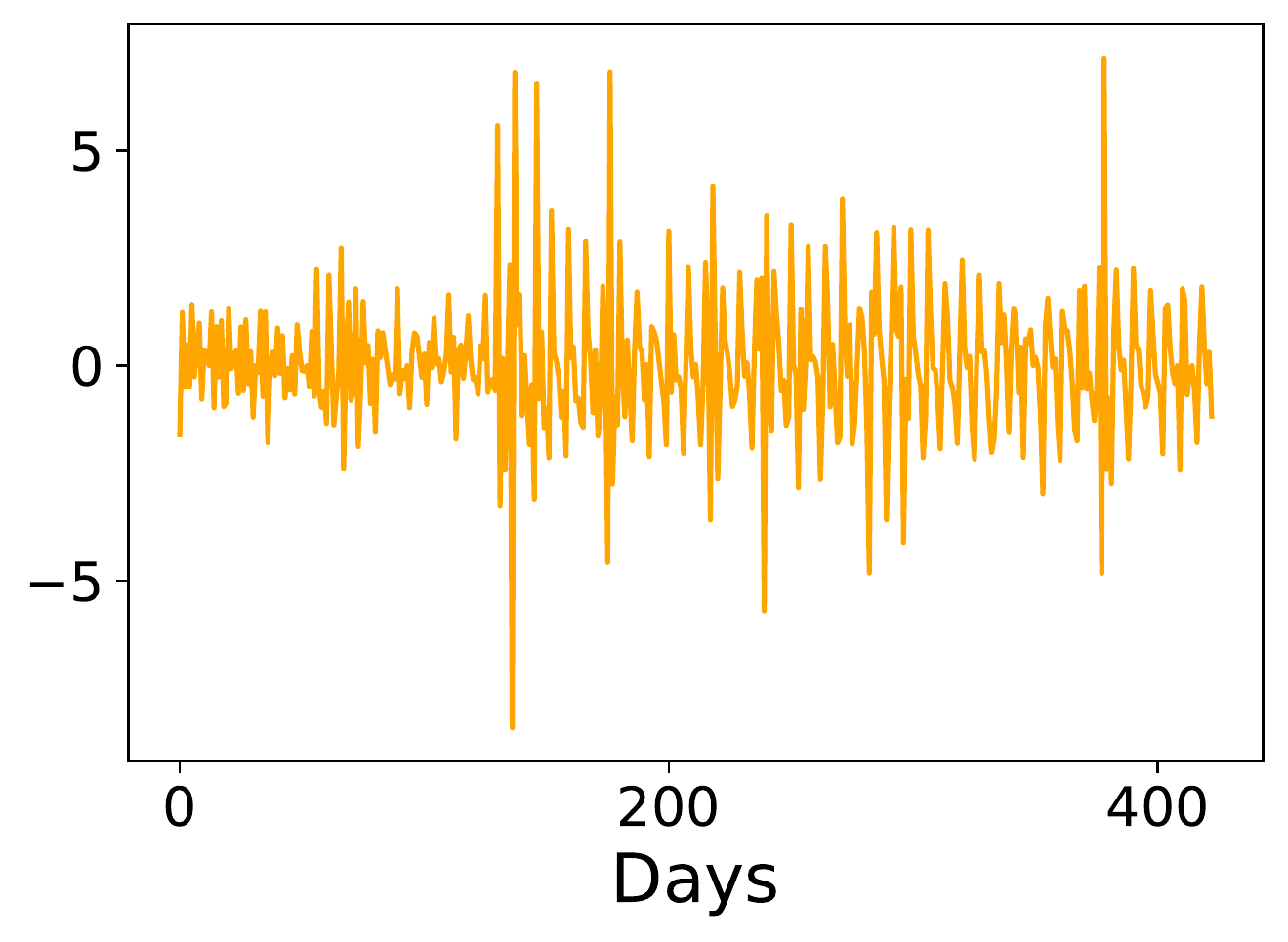}
    \end{minipage}        \\
Minnesota    &  \begin{minipage}{.15\textwidth}
      \includegraphics[width=1in]{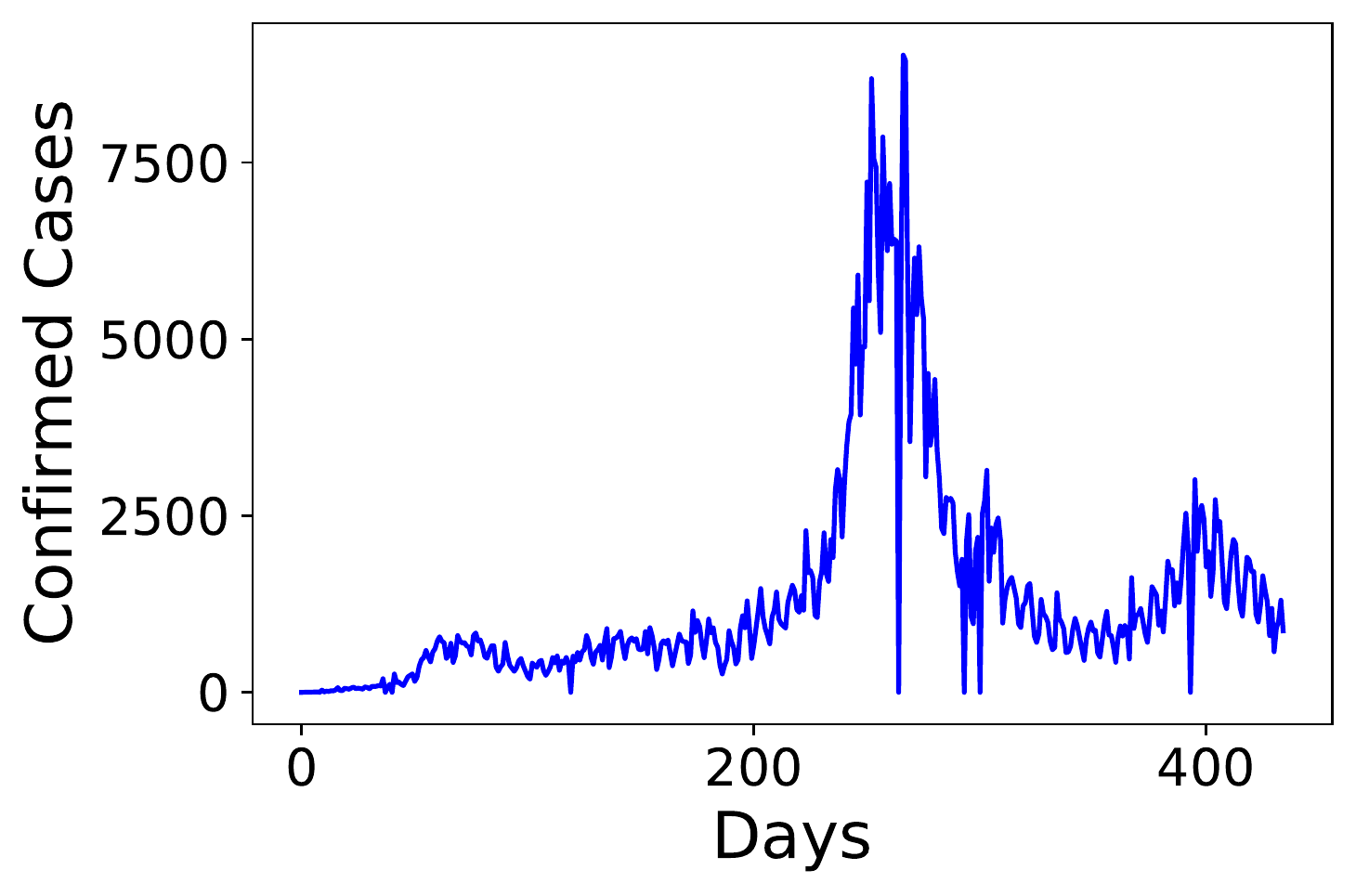}
    \end{minipage} & \begin{minipage}{.15\textwidth}
      \includegraphics[width=1in]{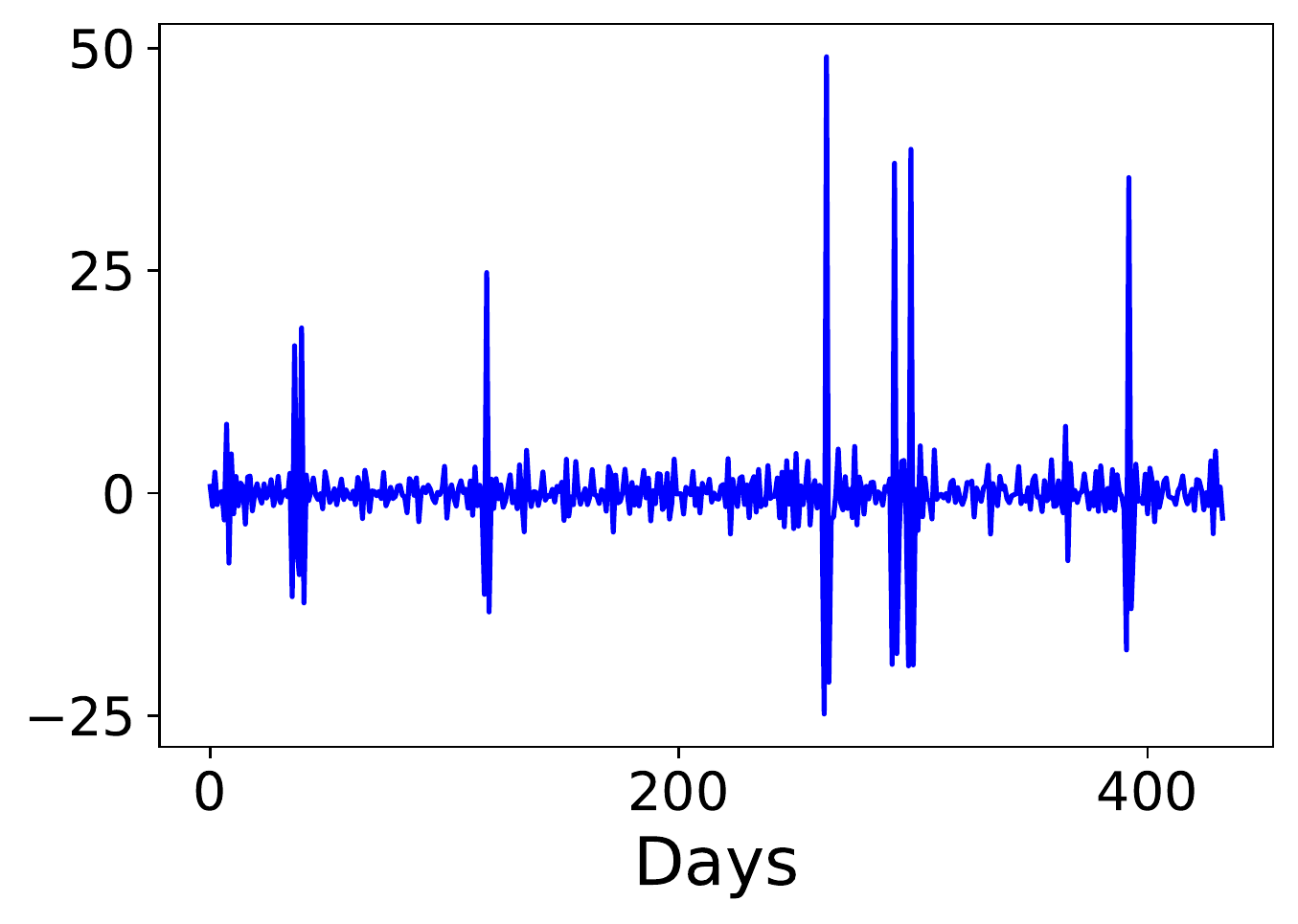}
    \end{minipage}& \begin{minipage}{.15\textwidth}
      \includegraphics[width=1in]{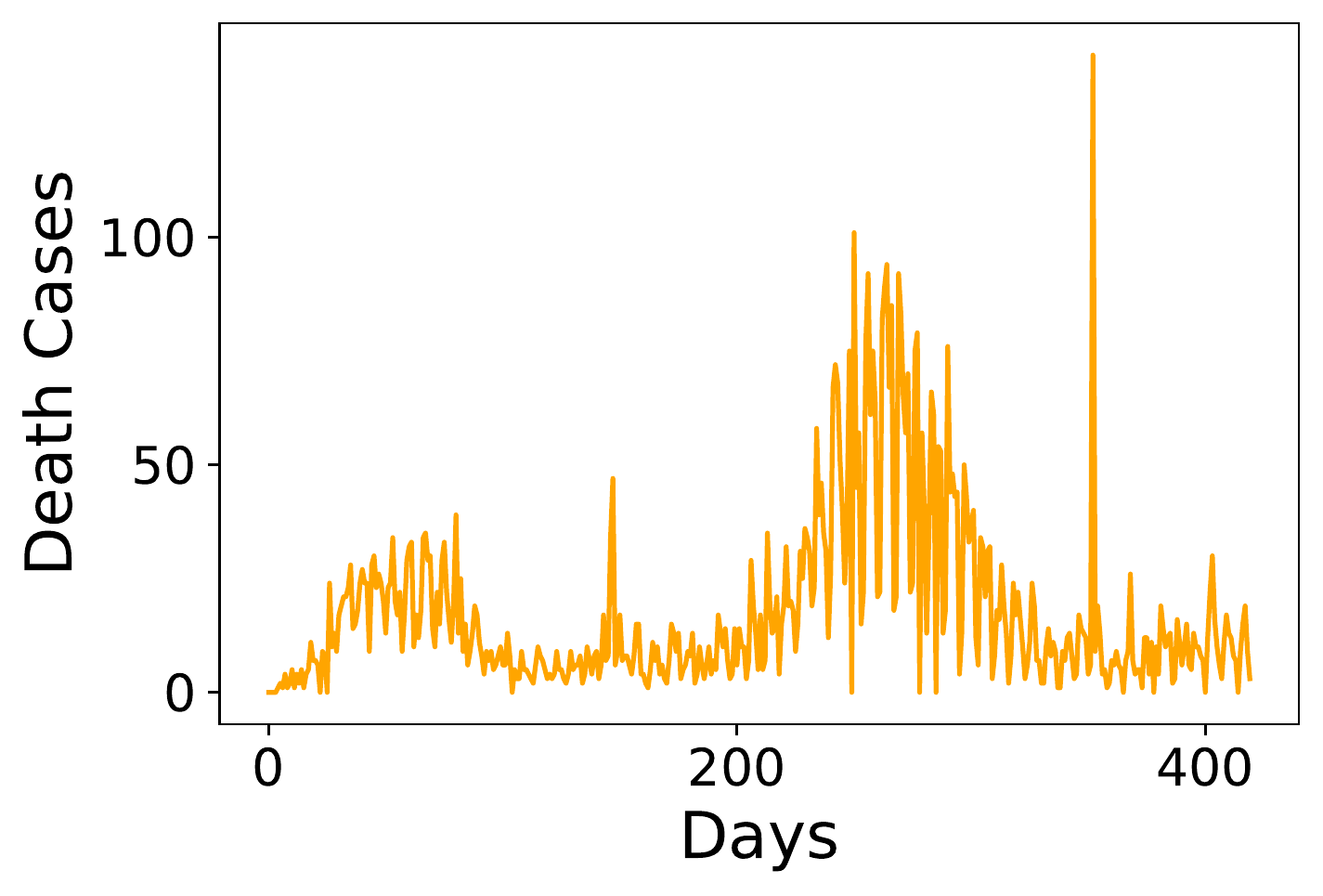}
    \end{minipage}  & \begin{minipage}{.15\textwidth}
      \includegraphics[width=1in]{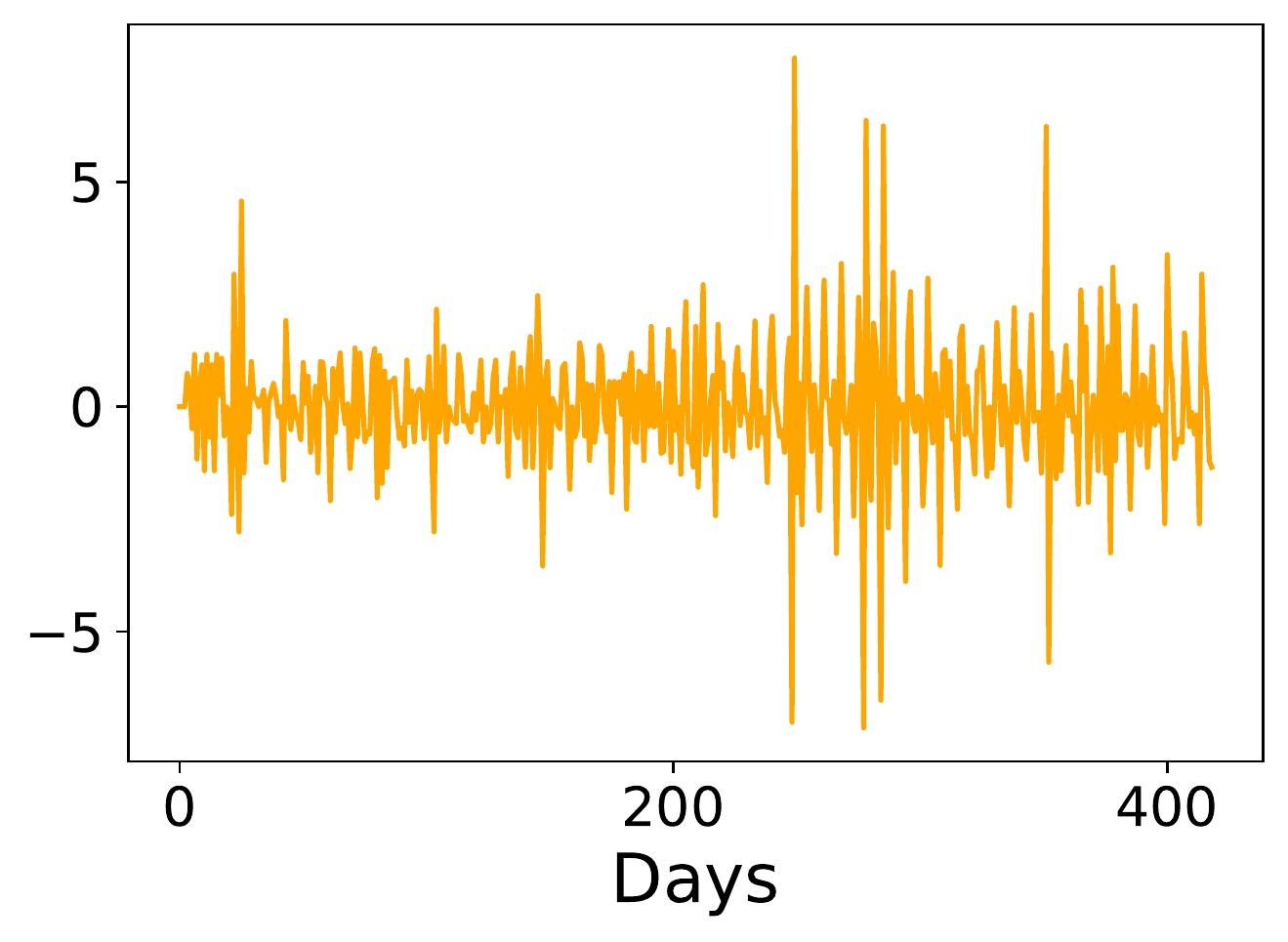}
    \end{minipage}  \\
Hawaii    &  \begin{minipage}{.15\textwidth}
      \includegraphics[width=1in]{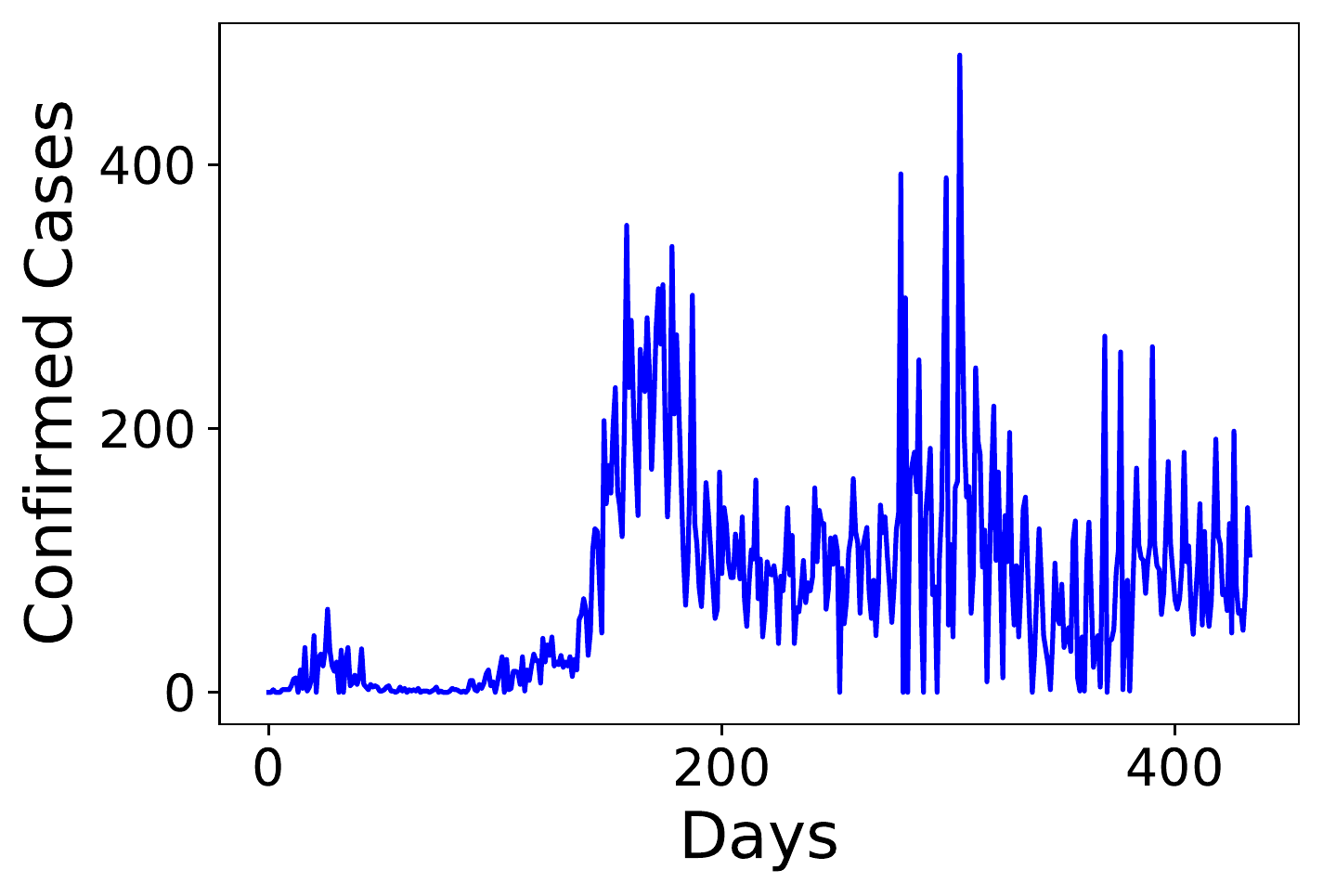}
    \end{minipage} & \begin{minipage}{.15\textwidth}
      \includegraphics[width=1in]{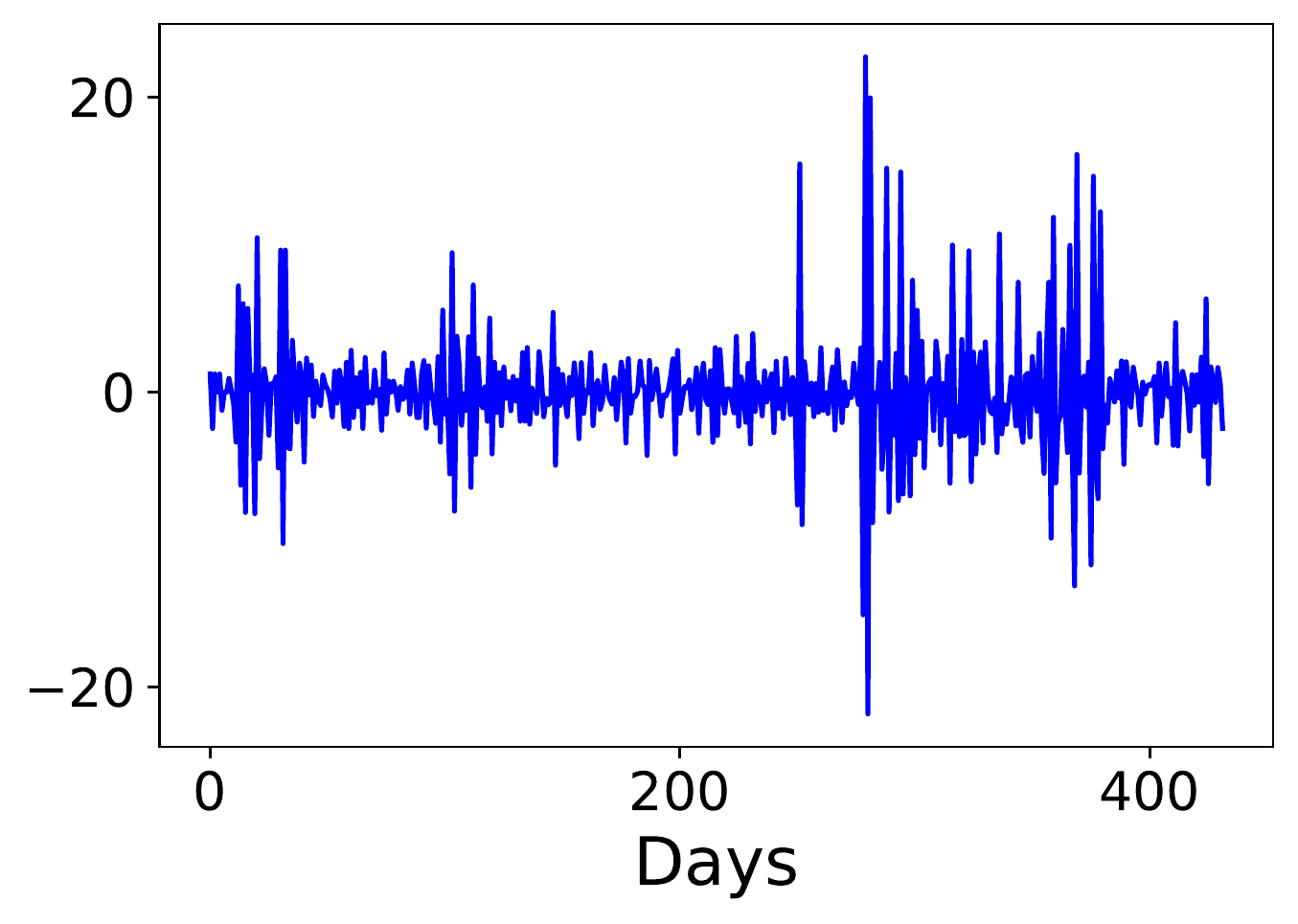}
    \end{minipage}& \begin{minipage}{.15\textwidth}
      \includegraphics[width=1in]{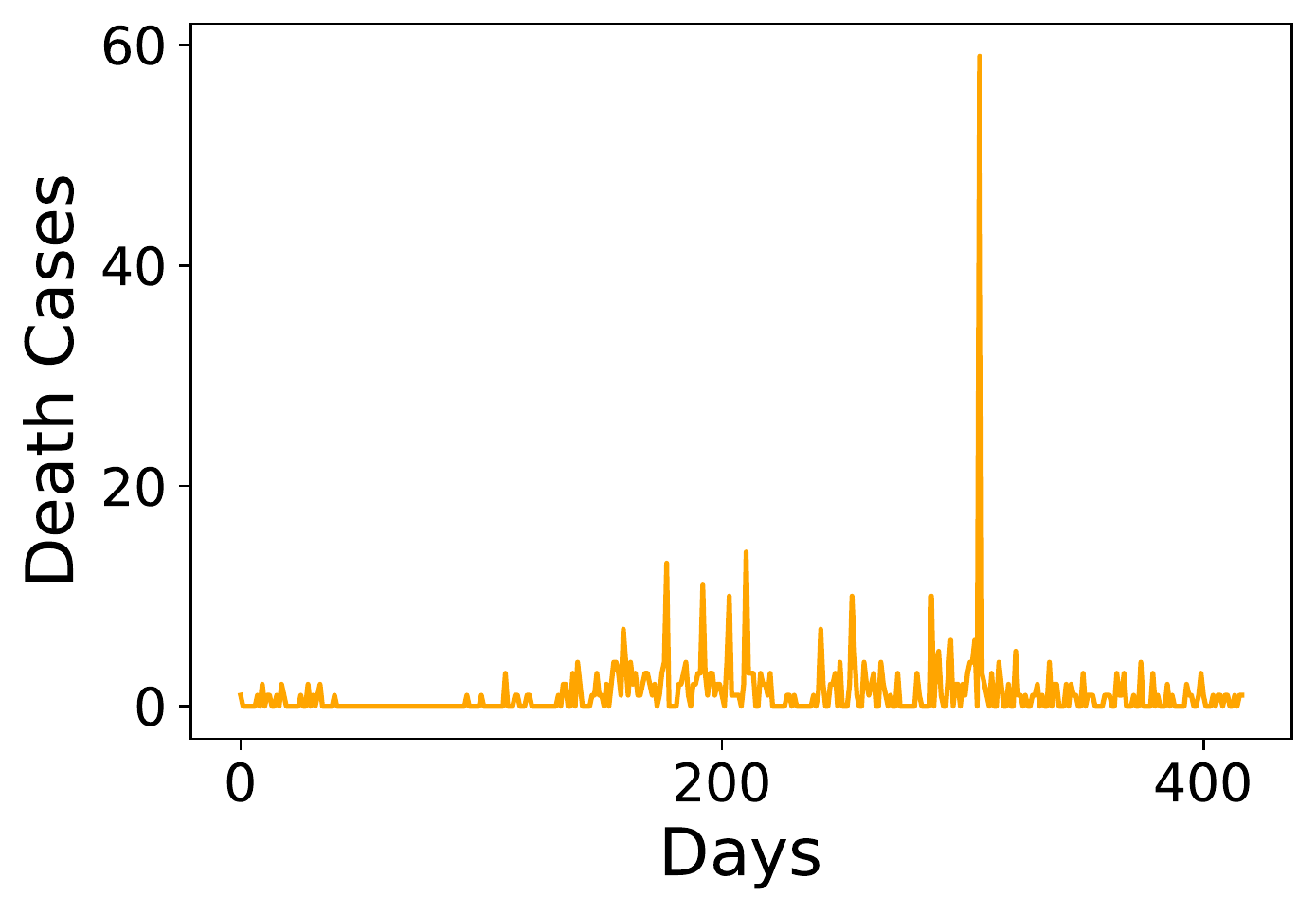}
    \end{minipage}  & \begin{minipage}{.15\textwidth}
      \includegraphics[width=1in]{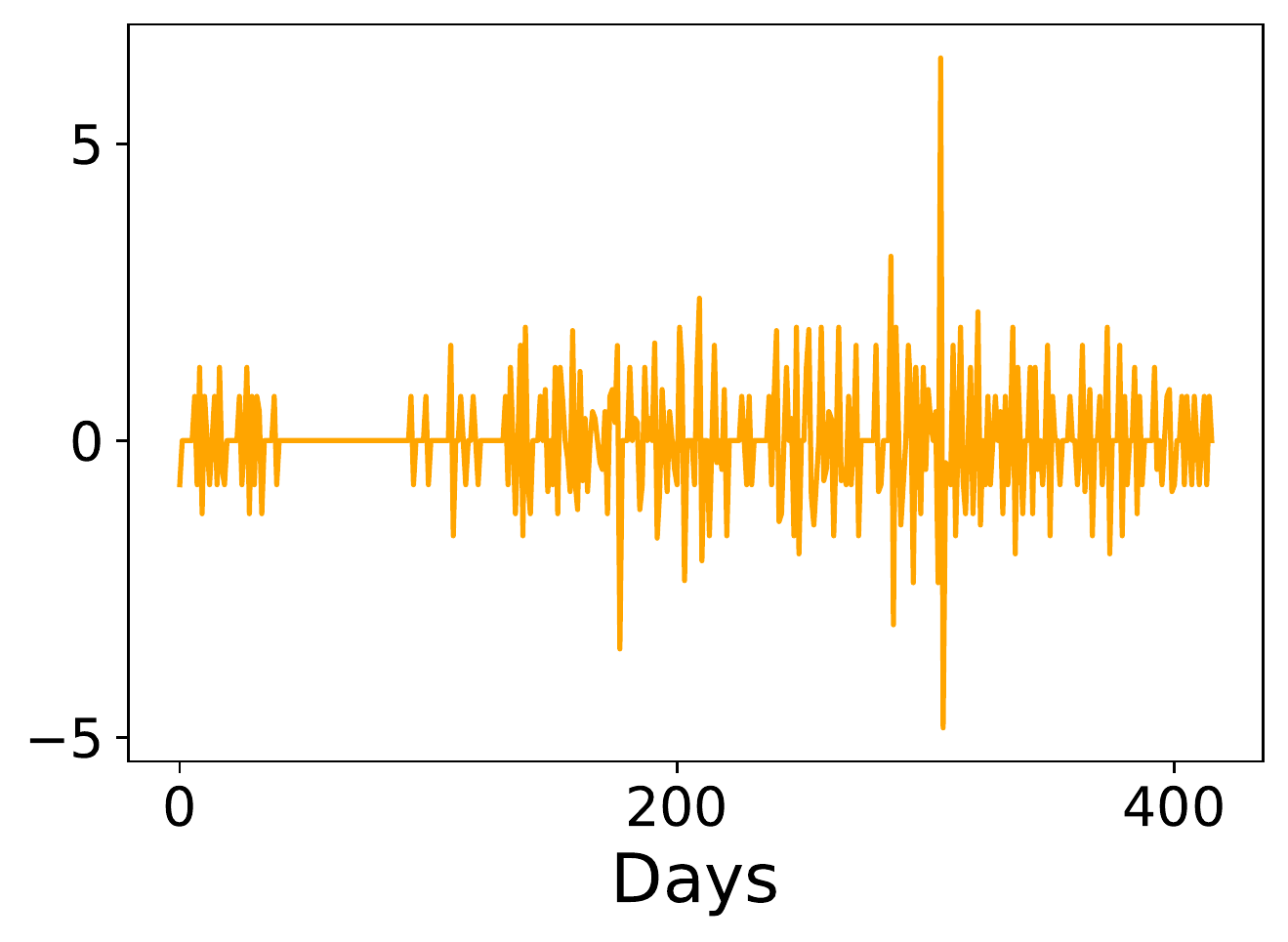}
    \end{minipage}  \\   
\Xhline{1.5pt}
\end{tabular}
\end{table*}

\begin{table*}[ht]
\centering
\caption{Results of the ADF test of processed data. Critical values for the test statistic at the 1\%, 5\%, and 10\% levels are -3.447, -2.869, and -2.571, respectively.}
\label{ADF}
\begin{tabular}{ccccccccccc}
\Xhline{1.5pt}
 & \multicolumn{2}{c}{\bfseries California} & \multicolumn{2}{c}{\bfseries New York} & \multicolumn{2}{c}{\bfseries Texas} & \multicolumn{2}{c}{\bfseries Minnesota} & \multicolumn{2}{c}{\bfseries Hawaii} \\
              & Confirmed & Death & Confirmed & Death & Confirmed & Death & Confirmed & Death & Confirmed & Death \\
ADF Statistic & -8.099  &  -3.576     & -8.770          &  -5.150     & -10.485          & -5.531      &    -10.352       &  -7.782     &   -12.023       &   -7.733    \\
p-value       & 1.311e-12  &  6.245e-3     &   2.541e-14        & 1.110e-5      &   1.187e-18       &  1.786e-6     &  2.527e-18         &  8.379e-12      &          2.991e-22  &  1.112e-11    \\
\Xhline{1.5pt}
\end{tabular}
\end{table*}

\begin{table}[H] 
\centering
\caption{Hyperparameter search space of SARIMA model.}
\label{hyperparameters_SARIMA}
\begin{tabular}{c c c c} 
\Xhline{1.5pt}
Hyperparameters     & Symbols & Hyperparameter Search Space \\
\hline
non-seasonal AR term & $p$ & [0, 1, 2, 3, 4] \\
seasonal AR term & $P$ & [0, 1, 2, 3, 4] \\
non-seasonal MA term & $q$ & [0, 1, 2, 3, 4] \\
seasonal MA term & $Q$ & [0, 1, 2, 3, 4]\\
non-seasonal differencing & $d$ & [1, 2] \\
seasonal differencing & $D$ & [0, 1, 2] \\
\Xhline{1.5pt}
\end{tabular}
\end{table}

\subsection{Experimental Procedures for SEIR-HCD}
For the SEIR-HCD model, we performed 7-day ahead and 28-day ahead forecasting of cumulative confirmed cases and death cases in CA, NY, TX, MN, and HI. The daily forecasting results on day $T$ were then obtained by taking the difference between cumulative cases on day $T$ and day $T-1$. We used $scipy.integrate.solve\_ivp$ in the $SciPy$ library to solve the set of ODE systems with initial conditions $S(0) = (N-n_{\inf})/N$, $I(0) = n_{\inf}/N$, and $E(0)=R(0)=H(0)=C(0)=D(0) = 0$, where $N$ represents the population size of each state and $n_{\inf}$ represents the number of infected people at $t = 0$. We took $n_{\inf} = 1$. The search ranges of the parameter estimation of the model are shown in Table~\ref{hyperparameters_SEIR_HCD}. The solutions were then used to fit to the training data. Furthermore, we assumed that the days that are closer to the prediction periods are more heavily weighted. We chose the period for optimization to be $21$ days. The L-BFGS-B algorithm was applied to minimize the mean squared logarithmic error function calculated from the above ODE system solutions.

\begin{table}[ht] 
\centering
\caption{Hyperparameter search space of SEIR-HCD model.}
\label{hyperparameters_SEIR_HCD}
\begin{tabular}{c c c} 
\Xhline{1.5pt}
Hyperparameters     & Symbols & Hyperparameter Search Space  \\ 
\hline
basic reproduction number & $R_0$ & [1.2, 3.6, 7.0] \\
average incubation period & $t_{\rm inc}$ & [4.0, 7.0, 14.0]  \\
average infectious period & $t_{\inf}$ & [2.9, 6.2, 10.1]  \\
average hospitalized period & $t_{\rm hosp}$ & [4, 12] \\
average critical period & $t_{\rm crt}$ & [5, 14] \\
ratio of asymptomatic infected & $\beta$ & [0.7, 0.9] \\
\makecell{ratio of hospitalised  \\ turning into critical state} & $\gamma$ & [0.1, 0.6] \\
\makecell{ratio of critical patients \\ resulting in death} & $\delta$ & [0.3, 0.8] \\
\Xhline{1.5pt}
\end{tabular}
\end{table}

\subsection{Experimental Procedures for ACTS}
All experiments with the ACTS model were implemented in Pytorch~\cite{paszke2019pytorch} on a single NVIDIA GTX 1080 Ti GPU (CUDA
10.2). We minimize the MAE loss for training. Each training epoch takes approximately 0.26 seconds. The hyperparameter search space for the model is listed in Table~\ref{hyperparameters_ACTS}.

\begin{table}[ht] 
\centering
\caption{Hyperparameter search space of the ACTS model.}
\label{hyperparameters_ACTS}
\begin{tabular}{c c c} 
\Xhline{1.5pt}
Hyperparameters     & Symbols & Hyperparameter Search Space  \\ 
\hline
$\sharp$ of training epochs & $N$ & [600, 1200, 1800] \\
hidden size & $d$ & [16, 32]  \\
learning rate & $\alpha$ & [0.001, 0.005, 0.01]  \\
\Xhline{1.5pt}
\end{tabular}
\end{table}

\subsection{Forecasting Performance Evaluation}
The forecasting performance of all three models was evaluated in terms of Accuracy, MAPE, WAPE, MAE, MSE, RMSE, and RMSLE, which are commonly used for time series forecasting in the literature~\cite{jin2020inter,zeroual2020deep,shoeibi2020automated}.

\section{Results} \label{results}
In this section, we quantify the impacts (performance improvement and variation) of each dimension (i.e. model selection, hyperparameter tuning, training TS length) on the predicted performance over our testing dataset (i.e. either keep the data in the
last 7 days or last 28 days of each state for validation), for the following $4$ prediction tasks across five different regions: 7-day ahead forecasts on confirmed cases (7-C), 28-day ahead forecasts on confirmed cases (28-C), 7-day ahead forecasts on death cases (7-D), and 28-day ahead forecasts on death cases (28-D).
First, we quantify the percentage that each dimension contributes to the performance improvement and variation in terms of Accuracy. Furthermore, we investigate the dimension that has the largest influence on the predictive performance. Finally, we analyze the relationship between performance improvement and variation through different evaluation indicators contributed by each dimension.

\subsection{Performance Improvement across Dimensions}
\begin{figure*}[htb]
    \centering
    \subfloat[7-day ahead forecasts on confirmed cases ]{\includegraphics[width=0.45\textwidth]{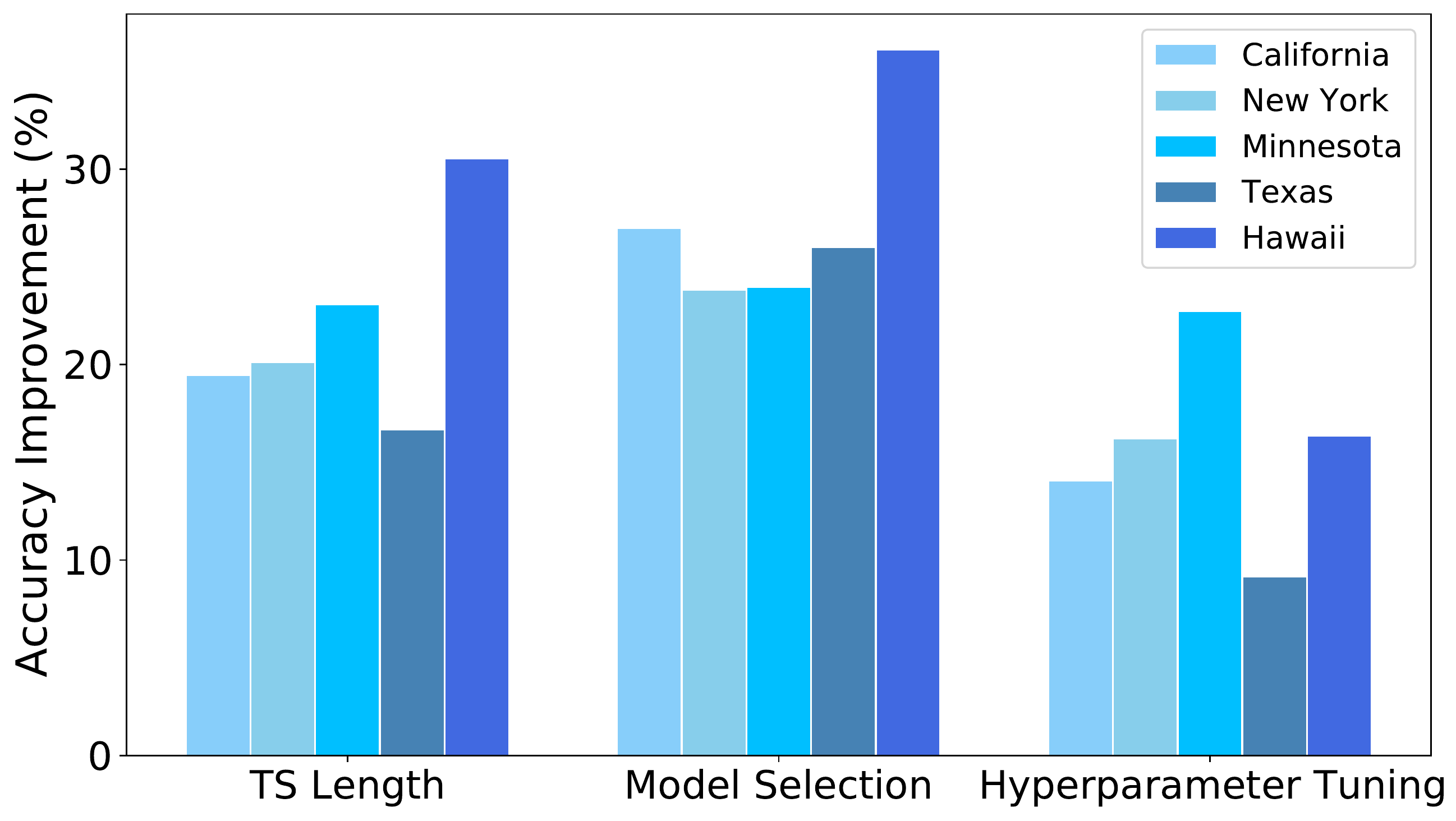}}
     \hspace{-0.0001mm}
    \subfloat[28-day ahead forecasts on confirmed cases]{\includegraphics[width=0.45\textwidth]{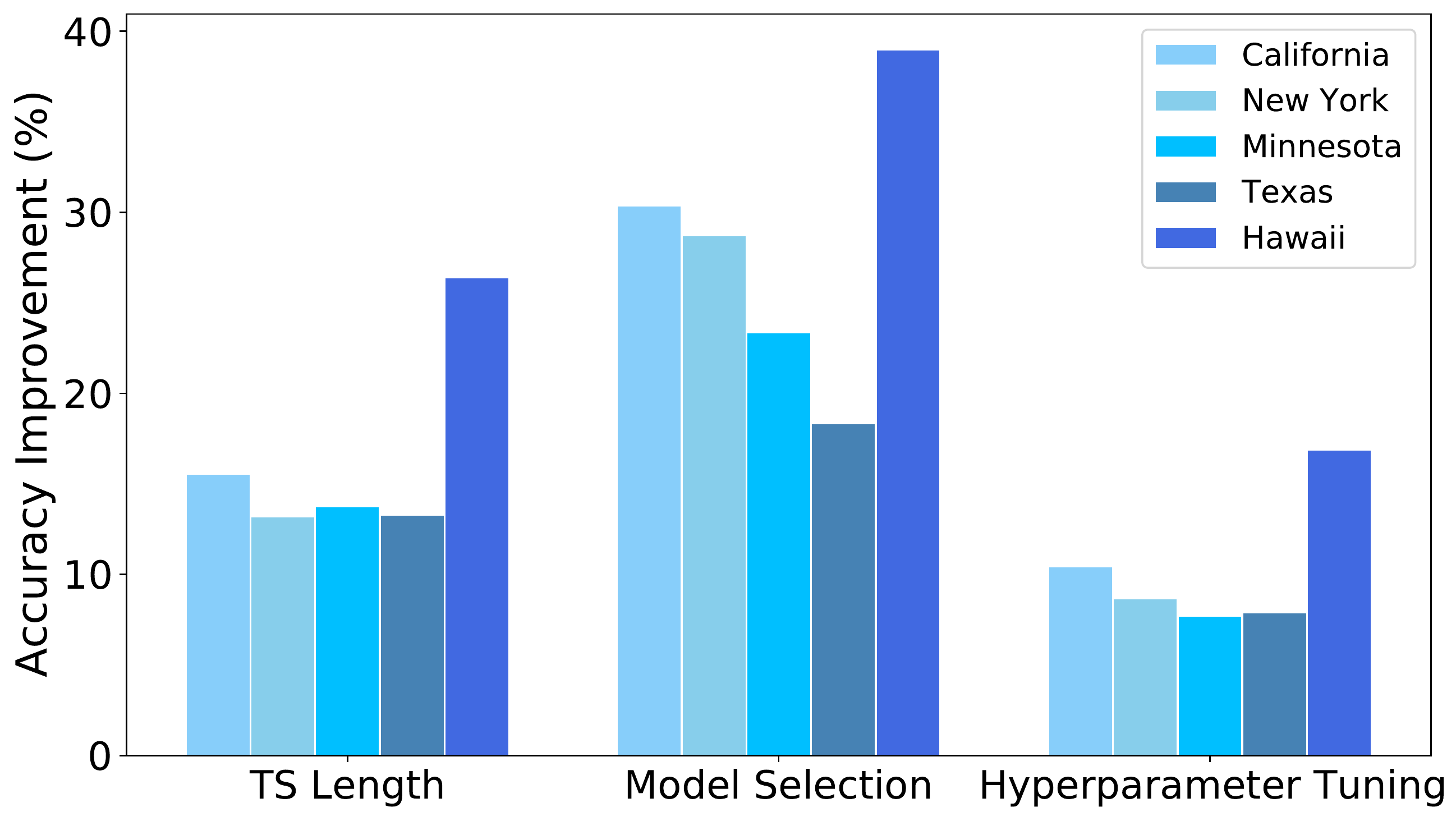}}
    \hfill
    \subfloat[7-day ahead forecasts on death cases ]{\includegraphics[width=0.45\textwidth]{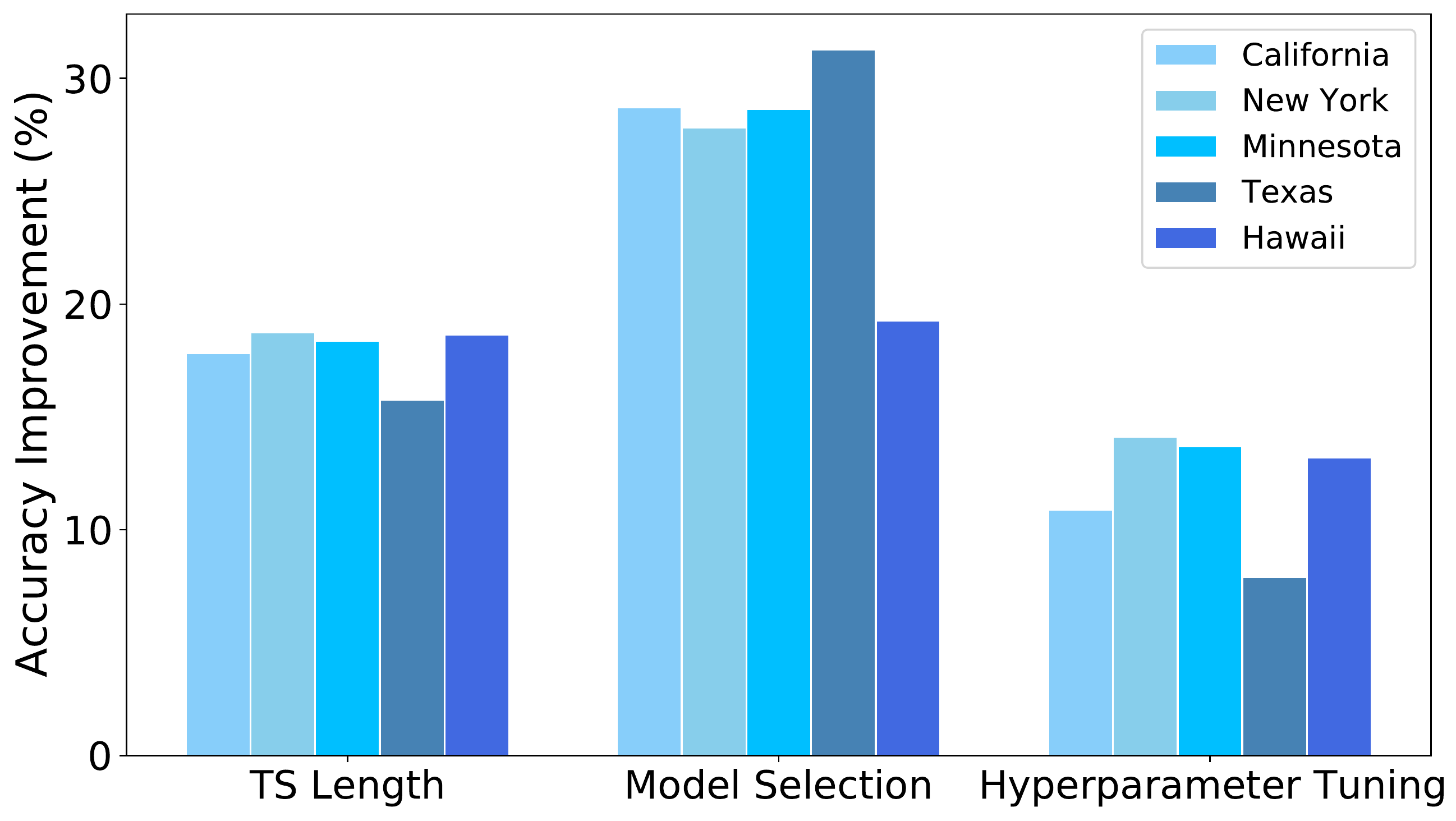}}
    \hspace{-0.0001mm}
    \subfloat[28-day ahead forecasts on death cases ]{\includegraphics[width=0.45\textwidth]{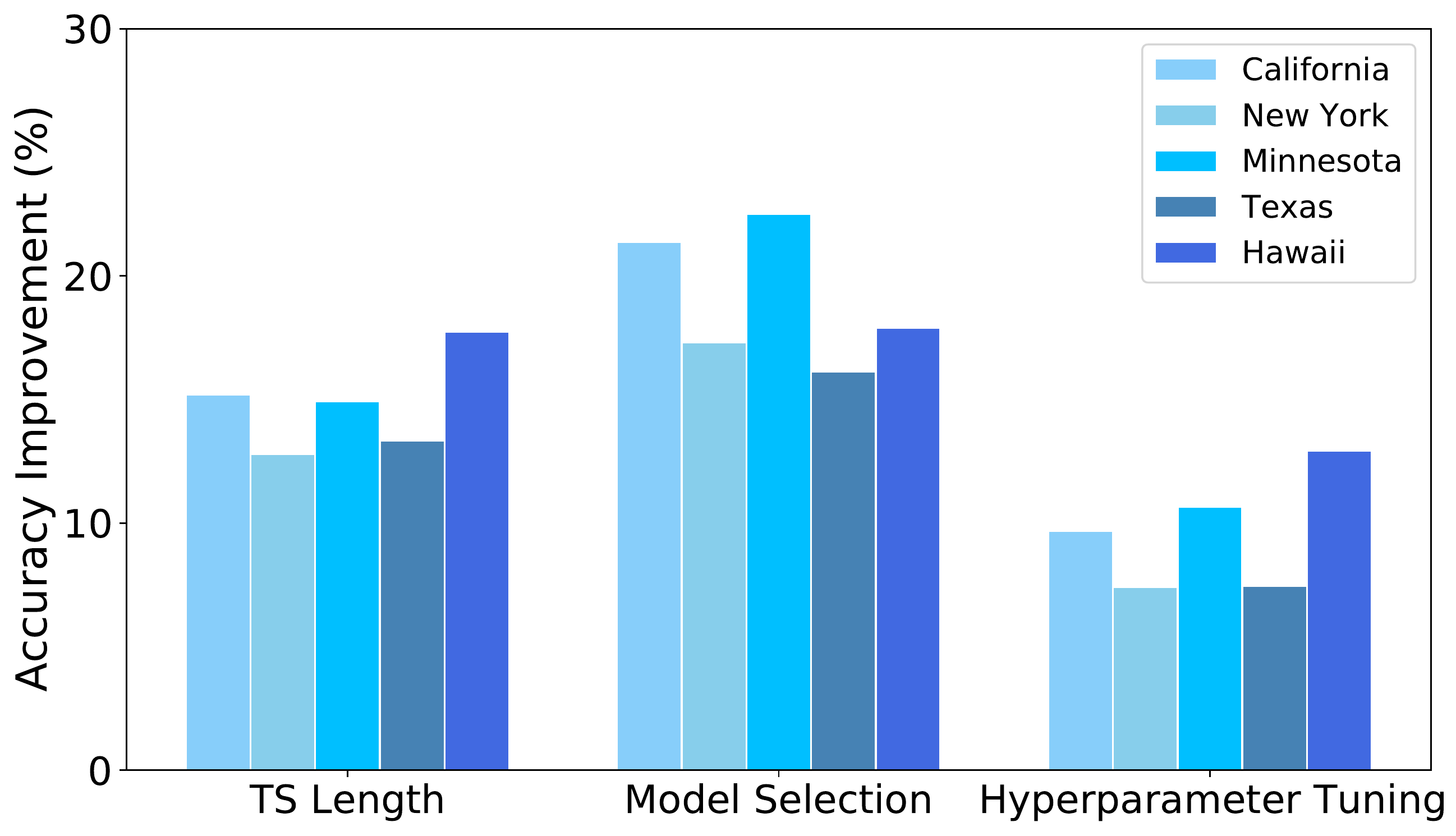}}
\caption{Performance improvement in the Accuracy score of each dimension over the baseline when tuning one dimension at a time while leaving others as baselines across five different states (CA, NY, MN, TX, HI). Model selection brings the greatest performance improvement, followed by TS length and hyperparameter tuning in decreasing order of improvement.}
\label{dim_imp}
\end{figure*}

For the hyperparameter tuning and training TS length across different regions, we define the $\textbf{\emph{baseline}}$ settings as the set of parameters that achieves the average performance for each model, and the length of TS = 200, respectively. For model selection, we choose SEIR-HCD, which exhibits the overall median performance among the three models. Then we quantify the performance improvement in terms of the Accuracy score of each dimension. For the SARIMA model, there are some combinations of the parameters that make MLE fail to converge. Hence, we remove all of the forecasting results for which some of the data was missing. In consideration of the ranges of the Accuracy from negative infinity to $1$ by definition, we disregard the scores with infinite values, and then we normalize the remaining valid Accuracy scores between $0$ and $1$. Fig.~\ref{dim_imp} shows the percentage that each dimension contributes to the improvement in the Accuracy score over baseline by tuning only one dimension at a time while leaving others at baseline settings for each prediction task. We  observe that model selection provides the largest performance gain (27.38\%, 27.96\%, 27.15\%, and 19.04\% of averaged performance improvement in the Accuracy score on the respective 7-C, 28-C, 7-D, and 28-D forecasting tasks across all regions), followed by TS length (21.98\%, 16.44\%, 17.87\%, and 14.79\% of averaged performance improvement in the Accuracy score on the respective 7-C, 28-C, 7-D, and 28-D forecasting tasks across all regions) and hyperparameter tuning (15.71\%, 10.33\%, 11.96\%, and 9.62\% of averaged performance improvement in the Accuracy score on the respective 7-C, 28-C, 7-D, and 28-D forecasting tasks across all regions) in decreasing order of improvement. 

To validate whether the Accuracy score is representative of performance across all the regions in all prediction tasks, we use $6$ other performance metrics described earlier to measure the degree of 
performance improvement of each dimension over all regions. We employ the same result processing techniques on other metrics as the Accuracy score by removing invalid forecasting results, and normalize the valid results. In addition to Accuracy, the performance improvement over all of the other metrics is defined as the reduced percent error between the tuned and the baseline settings. Table~\ref{metrics_imp} shows the averaged performance improvement of each dimension across all regions in different forecasting tasks. The results in Table~\ref{metrics_imp} indicate that model selection brings the biggest performance improvement regardless of the metrics we use, followed by training TS length and hyperparameter tuning. In other words, the relative contribution to performance improvement for each individual dimension based on other metrics is consistent with the Accuracy score, suggesting that the Accuracy score is a representative metric. 

\begin{table*}[htb]
\centering
\caption{Averaged performance improvement of each dimension on different metrics for $4$ forecasting tasks throughout all regions. MS refers to model selection, \\ HT refers to  hyperparameter tuning, and LEN refers to training TS length. The performance improvement of each dimension on other metrics displays \\ an order consistent with that of the Accuracy score.}
\label{metrics_imp}
\begin{tabular}{c|cccccccc}
\Xhline{1.5pt} 
\textbf{\makecell{Forecasting \\ Tasks}} &  & \textbf{Accuracy (\%)} & \textbf{MAE (\%)} & \textbf{MSE (\%)} & \textbf{RMSE (\%)} & \textbf{MAPE (\%)} & \textbf{WAPE (\%)} & \textbf{RMSLE (\%)}  \\
\hline
\multirow{3}{*}{\makecell{7-day ahead forecasts\\ on confirmed cases}}  & MS  & 27.38  & 29.07 & 23.61  & 28.52 & 29.30 & 29.07 &  29.90 \\ & HT  & 15.71 & 19.68  & 14.53 & 17.44 & 19.87 & 19.68  & 16.30  \\ & LEN & 21.98 & 21.68  & 14.98 & 19.74  & 22.05 & 21.68 & 28.50  \\
\hline                                          
\multirow{3}{*}{\makecell{28-day ahead forecasts\\ on confirmed cases}} & MS  & 27.96  & 28.43  & 23.41 & 28.84  & 28.77 & 28.43  & 30.43  \\ & HT  & 10.33 & 12.74  & 12.93 & 12.92 & 12.28 & 12.74  &  18.23 \\ & LEN & 16.44  & 15.54  & 14.06 & 15.46 & 15.55 & 15.51 & 26.81  \\
\hline
\multirow{3}{*}{\makecell{7-day ahead forecasts\\ on death cases}}      & MS  & 27.15 & 22.86 & 19.72 & 23.47 & 25.58 & 22.86 & 27.69  \\  & HT  & 11.96  & 12.97  & 11.74 & 13.08 & 13.25  & 12.97 & 19.69 \\ & LEN & 17.87 & 14.95 & 12.68 & 14.92 & 15.22 & 14.95 & 23.36  \\
\hline
\multirow{3}{*}{\makecell{28-day ahead forecasts\\ on death cases}}     & MS  & 19.04 & 20.70 & 17.83 & 21.41 & 22.47 & 20.70 & 22.64  \\ & HT  & 9.62 & 12.08  & 12.27  & 12.47  & 12.47  & 12.08  & 13.35 \\ & LEN & 14.79  & 13.49 & 14.72 & 13.48 & 13.69 & 13.49 & 20.54 \\
\Xhline{1.5pt} 
\end{tabular}
\end{table*}

\begin{figure*}[htb]
    \centering
    \subfloat[7-day ahead forecasts on confirmed cases ]{\includegraphics[width=0.45\textwidth]{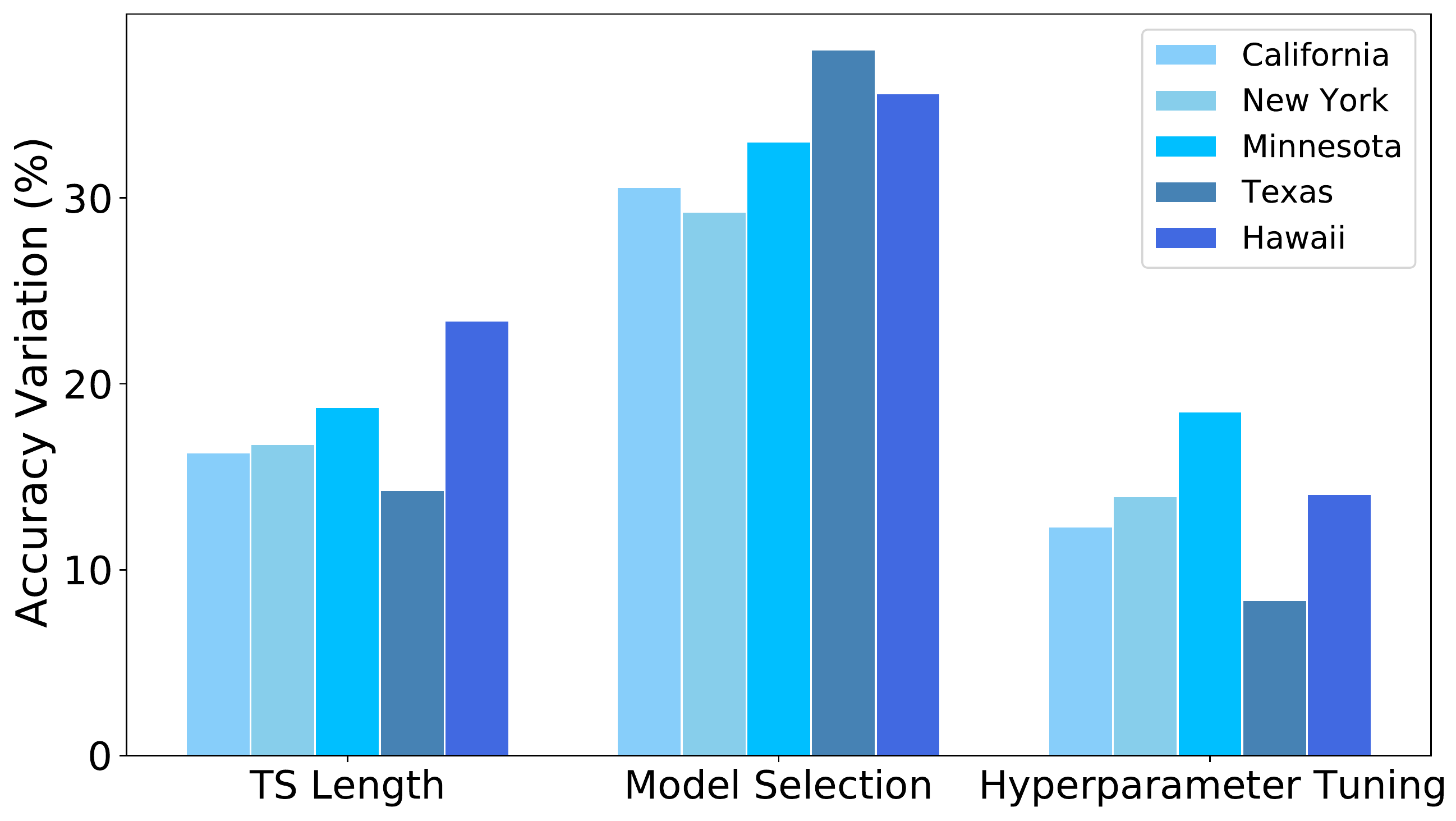}}
    \hspace{-0.0001mm}
    \subfloat[28-day ahead forecasts on confirmed cases]{\includegraphics[width=0.45\textwidth]{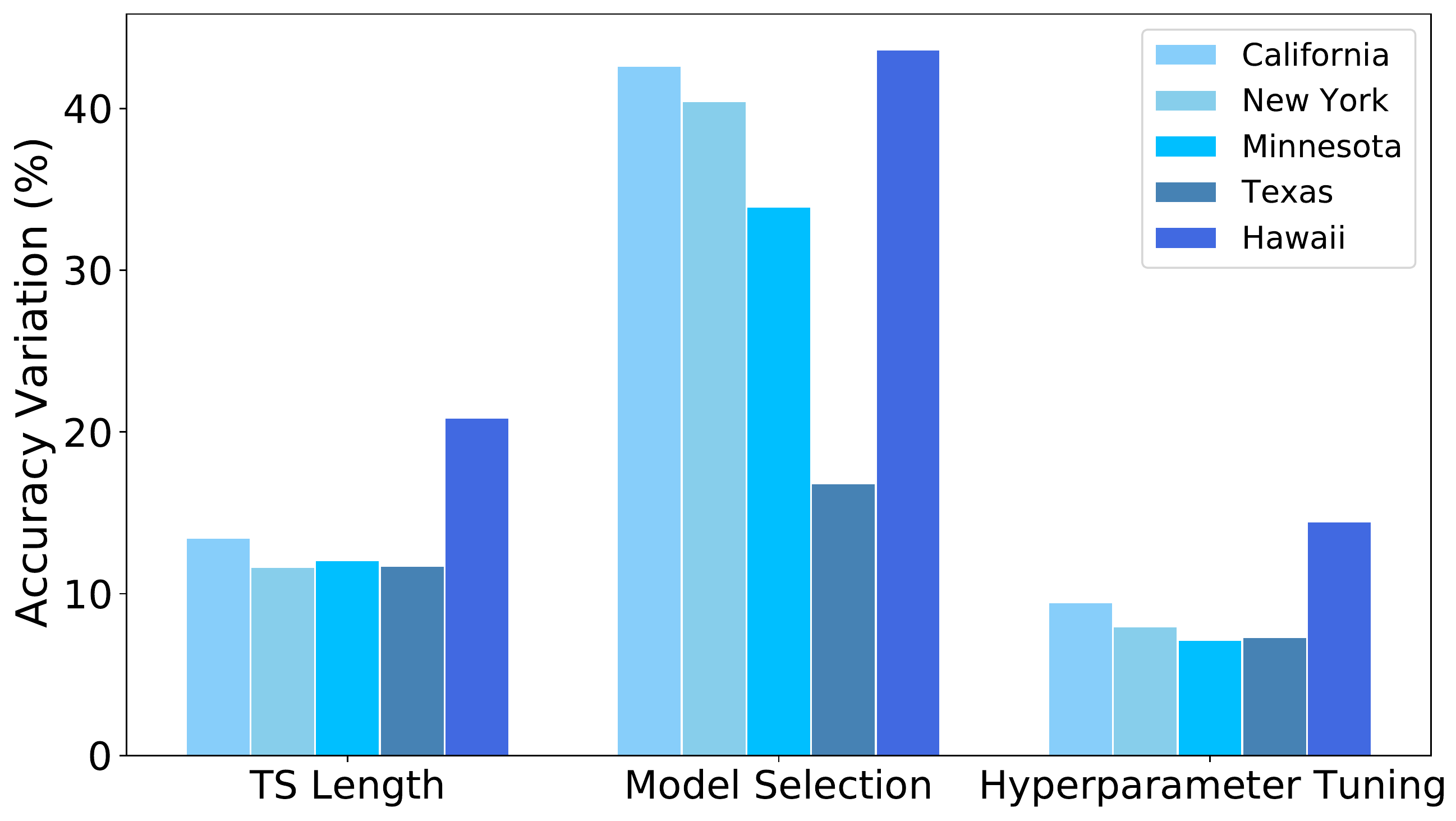}}
    \hfill
    \subfloat[7-day ahead forecasts on death cases ]{\includegraphics[width=0.45\textwidth]{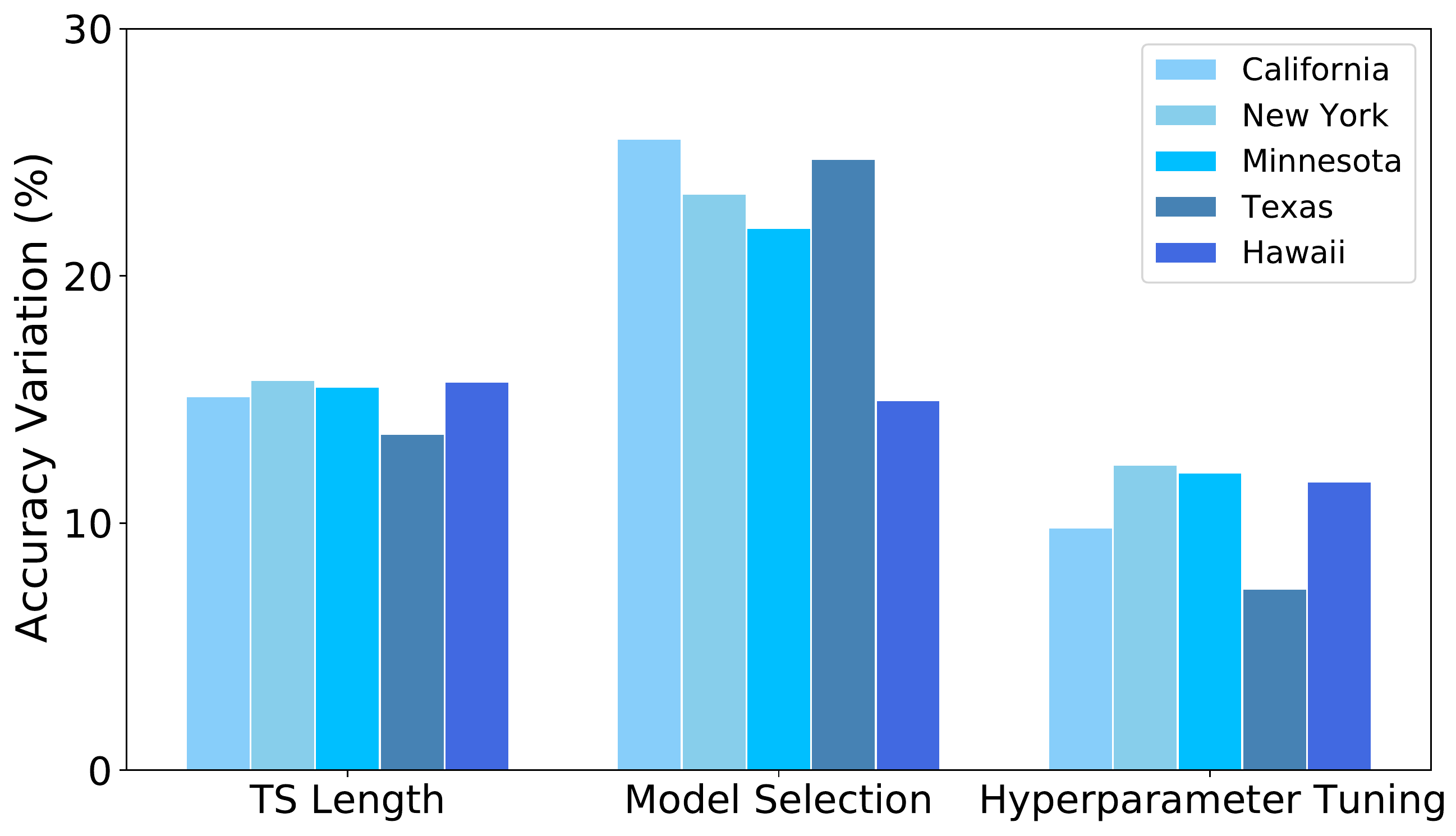}}
    \hspace{-0.0001mm}
    \subfloat[28-day ahead forecasts on death cases ]{\includegraphics[width=0.45\textwidth]{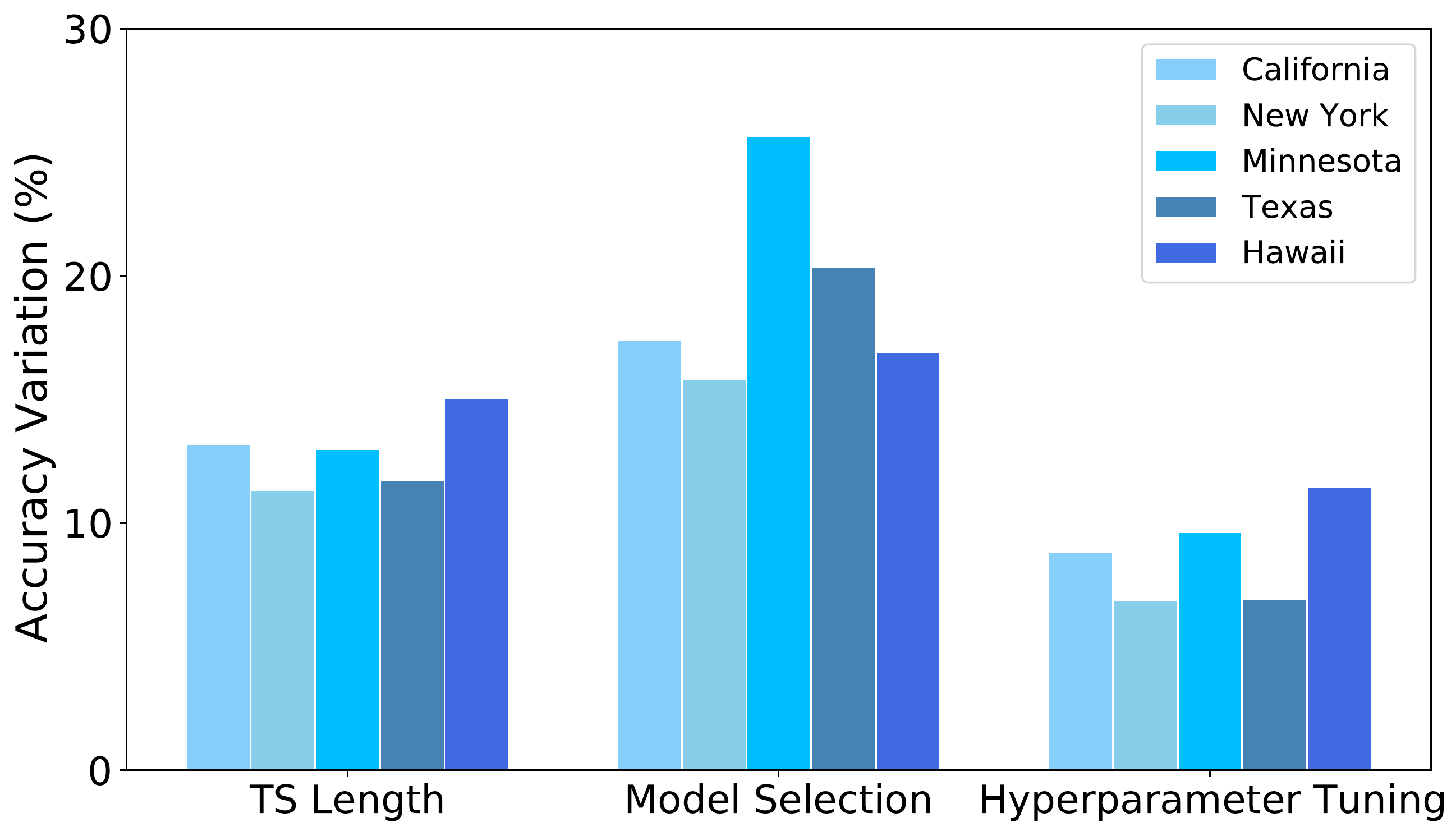}}
\caption{Performance variation in the Accuracy score of each dimension over the baseline when tuning only one dimension at a time while leaving others as baselines across five different states (CA, NY, MN, TX, HI). Model selection brings the greatest performance variation, followed by TS length and hyperparameter tuning in decreasing order of variation. Larger improvement also brings the risk of larger variation
for each dimension.}
\label{dim_var}
\end{figure*}

\begin{table*}[htbp]
\centering
\caption{Averaged performance variation of each dimension on different metrics for $4$ forecasting tasks. MS refers to model selection, HT refers to \\hyperparameter tuning, and LEN refers to training TS length. The performance variation of each dimension on other metrics displays an order \\ consistent with that of the Accuracy score.}
\label{metrics_var}
\begin{tabular}{c|cccccccc}
\Xhline{1.5pt} 
\textbf{\makecell{Forecasting \\ Tasks}} &  & \textbf{Accuracy (\%)} & \textbf{MAE (\%)} & \textbf{MSE (\%)} & \textbf{RMSE (\%)} & \textbf{MAPE (\%)} & \textbf{WAPE (\%)} & \textbf{RMSLE (\%)}  \\
\hline
\multirow{3}{*}{\makecell{7-day ahead forecasts\\ on confirmed cases}}  & MS & 27.38 & 35.07 & 27.95 & 33.40 & 35.67 & 35.07 & 35.33  \\ & HT  & 15.71 & 15.18 & 8.43 & 13.13 & 15.46 & 15.17 & 13.73  \\ & LEN & 21.98 & 18.99 & 11.90 & 17.47 & 19.30 & 18.99 & 23.45 \\
\hline                                          
\multirow{3}{*}{\makecell{28-day ahead forecasts\\ on confirmed cases}} & MS & 35.50 & 38.26 & 30.11 & 38.75 & 37.06 & 38.27 & 28.58 \\ & HT  & 9.28 & 8.78 & 7.27 & 8.88 & 8.94 & 8.78 & 14.62 \\ & LEN & 13.97 & 14.06 & 10.55 & 13.94 & 14.04 & 14.03 & 22.18  \\
\hline
\multirow{3}{*}{\makecell{7-day ahead forecasts\\ on death cases}}      & MS  & 22.10 & 28.43 & 22.49 & 29.69 & 32.68 & 28.42 & 37.29 \\  & HT  & 10.65 & 7.63 & 5.29 & 7.92 & 8.87 & 7.63 & 15.54 \\ & LEN & 15.15 & 14.24 & 11.50 & 14.20 & 14.49 & 14.24 & 22.98  \\
\hline
\multirow{3}{*}{\makecell{28-day ahead forecasts\\ on death cases}}     & MS  & 19.23 & 24.82 & 20.13 & 25.63 & 25.31 & 24.82 & 24.21  \\ & HT  & 8.75 & 5.83 & 5.03 & 6.28 & 7.21 & 5.83 & 10.36  \\ & LEN & 12.87 & 12.44 & 10.40 & 12.39 & 12.63 & 12.43 & 20.06\\
\Xhline{1.5pt} 
\end{tabular}
\end{table*}

\subsection{Performance Variation across Dimensions}
Next, we evaluate how much each dimension contributes to the performance variation in the Accuracy score. By tuning one dimension at a time while leaving others at baseline settings, we obtain a range of performance scores in terms of Accuracy. Performance variation is then defined as the difference between the maximum and the minimum score for each dimension. Higher variation in a single dimension implies that a poor choice in that dimension could produce remarkable performance loss.  Fig.~\ref{dim_var} shows the proportion of performance variation in the Accuracy score attributed to each individual dimension. We observe that model selection brings about the largest variation in performance (27.38\%, 35.50\%, 22.10\%, and 19.23\% of averaged performance variation in the Accuracy score on the respective 7-C, 28-C, 7-D, and 28-D forecasting tasks across all regions), followed by TS length (21.98\%, 13.97\%, 15.15\%, and 12.87\% of averaged performance variation in the Accuracy score on the respective 7-C, 28-C, 7-D, and 28-D forecasting tasks across all regions) and hyperparameter tuning (15.71\%, 9.28\%, 10.65\%, and 8.75\% of averaged performance variation in the Accuracy score on the respective 7-C, 28-C, 7-D, and 28-D forecasting tasks across all regions). The important takeaway here is that even though model selection is the largest contributor to performance improvement, if not carefully chosen, it can lead to larger performance degradation compared to other dimensions. 

To validate the representativeness of the accuracy score across all regions in different prediction tasks, $6$ other evaluation metrics are used to evaluate the performance variation on each dimension in consideration of all regions. Table~\ref{metrics_var} shows the averaged performance variation of each single dimension across all regions on different tasks. The results suggest that the proportion of performance variation in different metrics of each dimension follows an order that is consistent with the performance improvement in Table~\ref{metrics_imp}. Therefore, for different metrics, the dimension that brings about greater performance improvement may also contribute to larger variation in performance. For every step that researchers take when performing COVID-19 forecasting, they should always be aware of the trade-off between benefits (improvement in performance) and risks (variation in performance) when adjusting each dimension.

\section{Discussion} \label{discussion}
The COVID-19 pandemic is exponentially spreading around the world. Reliable forecasting on the number of confirmed and death cases provides pertinent information to decision-makers about the expected situations and the prevention measures that need to be taken. In consideration of the disparate impacts of social, economic, and environmental factors in different regions, we quantitatively analyze the predictive performance via disparate performance metrics in consideration of a wide range of configurations for COVID-19 case (confirmed and death) forecasting across different regions. Our study focuses on understanding the relationship between predictive performance and performance variation for each individual dimension in common time series forecasting tasks.

There are a few key takeaways from our results. First, for time series forecasting, if a dimension brings more performance improvement, it is also likely to bring greater performance degradation without good decisions in such dimension. Second, choosing the correct model is the most crucial step that
brings the largest impact in terms of the performance improvement and variation. 

Clearly, more forecasting models from diverse categories and different regions around the world should be considered to provide a more general conclusion. Other dimensions such as training time and robustness to incorrect input data, are worth further exploration.


\bibliographystyle{IEEEtran}
\bibliography{references}

\end{document}